\documentclass[nohyperref]{article}

\usepackage{microtype}
\usepackage{graphicx}
\usepackage{subfigure}
\usepackage{booktabs} % for professional tables
\usepackage{makecell}

\usepackage{hyperref}
\usepackage{multirow}
\usepackage[table,xcdraw]{xcolor}

\usepackage[accepted]{icml2023}

\usepackage{amsmath}
\usepackage{amssymb}
\usepackage{mathtools}
\usepackage{amsthm}

\usepackage[capitalize,noabbrev]{cleveref}

\theoremstyle{plain}

\theoremstyle{definition}
\newtheorem{definition}{Definition}

\theoremstyle{remark}
\newtheorem{remark}{Remark}

\usepackage[textsize=tiny]{todonotes}

\usepackage{amsmath,amsfonts,bm}

\def\Acal{{\mathcal{A}}}

\def\Mcal{{\mathcal{M}}}

\def\Pcal{{\mathcal{P}}}

\def\Scal{{\mathcal{S}}}
\def\Tcal{{\mathcal{T}}}

\def\eqref#1{equation~\ref{#1}}
\def\1{\bm{1}}

\DeclareMathAlphabet{\mathsfit}{\encodingdefault}{\sfdefault}{m}{sl}
\SetMathAlphabet{\mathsfit}{bold}{\encodingdefault}{\sfdefault}{bx}{n}

\newcommand{\E}{\mathbb{E}}

\usepackage[toc,page,header]{appendix}
\usepackage{minitoc}
\usepackage{wrapfig}

\begin{document}

\twocolumn[
\icmltitle{Constrained Decision Transformer for Offline Safe Reinforcement Learning}
% \icmltitlerunning{Submission and Formatting Instructions for ICML 2023}
% \icmltitle{Towards Adaptive Offline Safe Reinforcement Learning via Sequential Modeling}

% \icmltitle{Offline Safe Reinforcement Learning Beyond a Single Threshold}



\icmlsetsymbol{equal}{*}

\begin{icmlauthorlist}
\icmlauthor{Zuxin Liu}{equal,CMU}
\icmlauthor{Zijian Guo}{equal,CMU}
\icmlauthor{Yihang Yao}{CMU}
\icmlauthor{Zhepeng Cen}{CMU}
\icmlauthor{Wenhao Yu}{Google}
\icmlauthor{Tingnan Zhang}{Google}
\icmlauthor{Ding Zhao}{CMU}
\end{icmlauthorlist}

\icmlaffiliation{CMU}{Carnegie Mellon University}
\icmlaffiliation{Google}{Google Deepmind}

\icmlcorrespondingauthor{Zuxin Liu}{zuxinl@cmu.edu}

\icmlkeywords{Machine Learning, ICML}

\vskip 0.3in
]



\printAffiliationsAndNotice{\icmlEqualContribution} % otherwise use the standard text.

\begin{abstract}
Safe reinforcement learning (RL) trains a constraint satisfaction policy by interacting with the environment.
We aim to tackle a more challenging problem: learning a safe policy from an offline dataset.
We study the offline safe RL problem from a novel multi-objective optimization perspective and propose the $\epsilon$-reducible concept to characterize problem difficulties.
The inherent trade-offs between safety and task performance inspire us to propose the constrained decision transformer (CDT) approach, which can dynamically adjust the trade-offs during deployment.
Extensive experiments show the advantages of the proposed method in learning an adaptive, safe, robust, and high-reward policy.
CDT outperforms its variants and strong offline safe RL baselines by a large margin with the same hyperparameters across all tasks, while keeping the zero-shot adaptation capability to different constraint thresholds, making our approach more suitable for real-world RL under constraints.
The code is available at \url{https://github.com/liuzuxin/OSRL}
\end{abstract}

\doparttoc % Tell to minitoc to generate a toc for the parts
\faketableofcontents

\section{Introduction}
\label{sec:intro}
Learning high-reward policies from offline datasets has been a 
prevalent topic in reinforcement learning (RL) and has shown great promise in broad applications~\cite{fu2020d4rl, prudencio2022survey}.
Various learning paradigms are proposed to extract as much information as possible from pre-collected trajectories while preventing the policy from overfitting~\cite{ kostrikov2021offline, sinha2022s4rl}.
However, in the real world, many tasks can hardly be formulated by solely maximizing a scalar reward function, and the existence of various constraints restricts the domain of feasible solutions~\cite{gulcehre2020rl}.
For example, though numerous self-driving datasets are collected~\cite{sun2020scalability}, it is hard to define a single reward function to describe the task~\cite{lu2022imitation}.
The optimal driving policies should satisfy a set of constraints, such as traffic laws and physical dynamics.
Simply maximizing the reward may cause constraint violations and catastrophic consequences in safety-critical applications~\cite{chen2021context}.

Safe reinforcement learning aims to obtain a reward-maximizing policy within a constrained manifold~\cite{garcia2015comprehensive, brunke2021safe}, showing advantages to satisfy the safety requirements in real-world applications~\cite{ray2019benchmarking, gu2022review}.
However, most deep safe RL approaches focus on the safety during deployment, i.e., after training, while ignoring the constraint violation costs during training~\cite{xu2022trustworthy}.
The requirement of collecting online interaction samples brings challenges in ensuring training safety, because it is a non-trivial task to prevent the agent from executing unsafe behaviors during the learning process.
Though carefully designed correction systems or even human interventions can be used as a safety guard to filter unsafe action in training~\cite{saunders2017trial, dalal2018safe, wagener2021safe}, it could be expensive to be applied due to the low sample efficiency of many RL approaches~\cite{xie2021policy}. 

This paper studies the problem of learning constrained policies from offline datasets such that the safety requirements can be met both in training and deployment.
Several recent works tackle the problem by bridging the ideas in offline RL and safe RL domains, such as using pessimistic estimations~\cite{xu2022constraints} or the stationary distribution correction technique~\cite{liu2020constrained, lee2022coptidice}.
A constrained optimization formulation and Lagrange multipliers are usually adopted when updating the policy, targeting to find the most rewarding policy while satisfying the constraints~\cite{le2019batch}.
However, these approaches require setting a constant constraint threshold before training, and thus the trained agents can not be adapted to other constraint conditions.
We believe \textit{the capability of adapting the trained policy to different constraint thresholds is important} for many practical applications, because imposing stricter constraints is usually at the cost of sacrificing the task performance and inducing conservative behaviors~\cite{liu2022robustness}.
Therefore, we aim to study a training scheme such that the trained agent can dynamically adjust its constraint threshold, such that we can control its deployment conservativeness without further fine-tuning or re-training.

We also observe that the taxonomy of offline safe RL datasets is not adequately discussed in the literature, while we believe the characterization of a dataset can significantly influence the problem difficulty.
We provide a novel view of the offline safe RL problem using tools from the multi-objective optimization (MOO) domain, which unveils the inherent trade-off between safety performance and task reward.
The trade-offs can be described by a function with respect to the dataset and the constraint threshold, which inspires us to propose the Constrained Decision Transformer (CDT) approach.
CDT leverages the return-conditioned sequential modeling framework \cite{chen2021decision} to achieve zero-shot adaptation to different constraint thresholds at deployment while maintaining safety and high reward. 
The main contributions are summarized as follows:

\begin{itemize}
    \item We study the offline safe RL problem beyond a single pre-defined constraint threshold from a novel MOO perspective. The insights suggest the limitations of existing offline safe RL training paradigms and motivate us to propose CDT by leveraging the return-conditioned sequential modeling capability of Transformer.
    \item We propose three key techniques in CDT that are important in learning an adaptive and safe policy.
    To the best of our knowledge, CDT is the first successful offline safe RL approach that can achieve zero-shot adaptation to different safety requirements after training, without solving a constrained optimization.
    \item Extensive experiments show that CDT outperforms the baselines and its variants in terms of both safety and task performance by a large margin. CDT can generalize to different cost thresholds without re-training the policy, while all the prior methods fail.
\end{itemize}

\vspace{-1mm}
\section{Related Work}
\vspace{-1mm}
\label{sec:related}
\textbf{Safe RL.} Constrained optimization techniques are usually adopted to solve safe RL problems~\cite{garcia2015comprehensive, sootla2022saute, yang2021wcsac, flet2022saac, ji2023omnisafe}. Lagrangian-based methods use a multiplier to penalize constraint violations~\cite{chow2017risk, tessler2018reward, stooke2020responsive, chen2021primal}. 
Correction-based approaches project unsafe actions to the safe set, aiming to incorporate domain knowledge of the problem to achieve safe exploration~\cite{zhao2021model, luo2021learning}. 
Another line of work performs policy optimization on surrogate policy spaces via low-order Taylor approximations \cite{achiam2017constrained, yang2020projection} or variational inference \cite{liu2022constrained}.
However, ensuring zero constraint violations during training is still a challenging problem.

% \cite{ray2019benchmarking}.

\textbf{Offline RL.}
Offline RL targets learning policies from collected data without further interaction with the environment~\cite{ernst2005tree, tarasov2022corl}.
% One main challenge for offline RL is state-action distribution shift~\citep{levine2020offline, prudencio2022survey}. 
Many regularization methods are proposed to address the state-action distribution shift problem between the static dataset and physical world~\citep{levine2020offline, prudencio2022survey}.
One type of approach limits the discrepancy between learned policy and behavioral policy \citep{fujimoto2019off, kumar2019stabilizing, peng2019advantage, nair2020awac, fujimoto2021minimalist}. 
Another way is to use value regularization as implicit constraints\citep{wang2020critic}, e.g., optimizing the policy based on a conservative value estimation~\citep{kumar2020conservative}. 
In addition to the above pessimism mechanism, stationary distribution correction (DICE)-style methods train the policy by importance sampling, which reduces the estimation variance~\citep{nachum2019algaedice, nachum2019dualdice, zhang2020gendice}. 
Recent research also shows the great success of leveraging the power of Transformer to perform behavior cloning style policy optimization~\citep{janner2021offline, chen2021decision, furuta2022generalized}.

\textbf{Offline RL with safety constraints.} Several recent works study the offline safe RL problem, aiming to achieve zero constraint violations during training \cite{le2019batch}. They utilize the ideas from both offline RL and safe RL, such as using the DICE-style technique to formulate the constrained optimization problem \cite{polosky2022constrained, lee2022coptidice}.
Lagrangian-based approaches are also explored due to their simplicity of combining with existing offline RL methods, and are shown to be effective when using conservative cost estimation~\cite{xu2022constraints}.
However, how to adapt a trained safe policy to various constraint thresholds is rarely discussed in the literature.

% \textbf{Multi-objective RL.}

\vspace{-1mm}
\section{Preliminaries}
\vspace{-1mm}
\label{sec:preliminary}
\subsection{CMDP and Safe RL}

Safe RL can be described under the Constrained Markov Decision Process (CMDP) framework~\cite{altman1998constrained}.
A finite horizon CMDP $\Mcal$ is defined by the tuple $(\Scal, \Acal, \Pcal, r, c, \mu_0)$, where $\Scal$ is the state space, $\Acal$ is the action space, $\Pcal:\Scal \times \Acal \times \Scal \xrightarrow{} [0, 1]$ 
is the transition function, $r:\Scal \times \Acal \times \Scal \xrightarrow{} \mathbb{R}$ is the reward function, and $\mu_0: \Scal \xrightarrow[]{} [0,1]$ is the initial state distribution.
CMDP augments MDP with an additional element $c:\Scal \times \Acal \times \Scal \xrightarrow{} [0, C_{max}]$ to characterize the cost for violating the constraint, where $C_{max}$ is the maximum cost.
Note that this work can be directly applied to multiple constraints and partially observable settings, but we use CMDP with a single constraint for ease of demonstration.

A safe RL problem is specified by a CMDP and a constraint threshold $\kappa \xrightarrow[]{} [0, +\infty)$.
Let $\pi:\Scal \times \Acal\rightarrow [0,1]$ denote the policy and $\tau = \{s_1, a_1, r_1, c_1 ..., s_T, a_T, r_T, c_T \}$ denote the trajectory, where $T = |\tau|$ is the maximum episode length.
We denote $R(\tau) = \sum_{t=1}^{T} r_t$ as the reward return of the trajectory $\tau$ and $C(\tau)=\sum_{t=1}^{T} c_t$ as the cost return.
% The reward return and cost return of a trajectory at timestep $t$ are denoted as $V_r^\tau(s_t) = \sum_{t'=t}^T r_{t'}$ and $V_c^\tau(s_t) = \sum_{t'=t}^T c_{t'}$, respectively.
The goal of safe RL is to find the policy that maximizes the reward return while limiting the cost incurred from constraint violations to the threshold $\kappa$:
\begin{equation}
\vspace{-2mm}
   \max_{\pi} \mathbb{E}_{\tau \sim \pi}  \big[R(\tau)], \quad s.t. \quad \mathbb{E}_{\tau \sim \pi} \big[C(\tau)] \leq \kappa. 
   \label{eq:safe-rl}
\end{equation} 
In the offline setting, the agent can not collect more data by interaction but only access pre-collected trajectories from arbitrary and unknown policies, which brings challenges to solving this constrained optimization problem.

\subsection{Decision Transformer for Offline RL}
\label{sec:dt}
Decision Transformer (DT)~\cite{chen2021decision} is a type of sequential modeling technique to solve offline RL problems, without considering the constraint in Eq.~(\ref{eq:safe-rl}).
Unlike classical offline RL approaches that parametrize a single state-conditioned policy $\pi(a|s)$, DT takes in a sequence of reward returns, states, and actions as input tokens, and outputs the same length of predicted actions.
Given a trajectory $\tau$ of length $T$, the reward return at timestep $t$ is computed by $R_t = \sum_{t'=t}^{T} r_{t'}$, then we obtain 3 types of tokens for DT: reward returns $\mathbf{R}=\{R_1, ..., R_T\}$, states $\mathbf{s}=\{s_1,...,s_T\}$, and actions $\mathbf{a} = \{a_1, ..., a_T\}$.
The input sequence for DT at timestep $t$ is specified by a context length $K\in\{1, ..., t-1\}$, and the tokens are 
$\mathbf{R}_{-K:t}=\{R_K, ..., R_t\}$, $\mathbf{s}_{-K:t}=\{s_K, ..., s_t\}$ and $\mathbf{a}_{-K:t-1}=\{a_K, ..., a_{t-1}\}$.
The DT policy is parametrized by the GPT architecture~\cite{radford2018improving} with a causal self-attention mask, such that the action sequences are generated in an autoregressive manner. 
Namely, DT generates a deterministic action at timestep $t$ by $\hat{a}_t = \pi_{\text{DT}}(\mathbf{R}_{-K:t}, \mathbf{s}_{-K:t}, \mathbf{a}_{-K:t-1})$.
Then the policy can be trained by minimizing the loss between the predicted actions and the ground-truth actions in a sampled batch of data.
Typically, DT uses the cross-entropy loss for discrete action spaces and the $\ell_2$ loss for continuous action spaces.

\vspace{-1mm}
\section{Method}
\vspace{-1mm}
\label{sec:method}
\subsection{The Offline Safe RL Problem}
\label{sec:offline-safe-rl}

In this section, we revisit the offline safe RL problem and investigate its taxonomy based on collected datasets' cost threshold and properties.
Denote $\Tcal = \{\tau_1, \tau_2,...\}$ as a dataset of trajectories.
For the sake of subsequent analysis, we make the assumption that the dataset is both \textit{clean} and \textit{reproducible}, meaning that any trajectory in the dataset can be reliably reproduced by a policy. 
This is an important precondition, as characterizing noisy datasets that contain outliers in highly stochastic environments is challenging and lies beyond the discussion scope of this paper.

To describe a dataset $\mathcal{T}$ with both reward and cost metrics, we introduce the Pareto Frontier (PF), Inverse Pareto Frontier (IPF), and the Reward Frontier (RF) functions that are inspired by the MOO domain.
The PF of a dataset $\Tcal$  is computed by the maximum reward of trajectories under cost threshold $\kappa \in [0,\infty)$:
$$\text{PF}(\kappa, \Tcal) = \max_{\tau \in \mathcal{T}} R(\tau), \quad s.t. \quad C(\tau) \leq \kappa.$$
Similarly, the IPF of a dataset $\Tcal$ is defined by the maximum reward beyond cost threshold $\kappa \in [0,\infty)$:
$$\text{IPF}(\kappa, \Tcal) = \max_{\tau \in \mathcal{T}} R(\tau), \quad s.t. \quad C(\tau) \geq \kappa.$$
The RF is defined by the maximum reward with cost $\kappa \in \mathbb{C}$, where $\mathbb{C} := \{C(\tau): \tau \in \mathcal{T}\}$ is the set of all the possible episodic cost in $\mathcal{T}$:
$$\text{RF}(\kappa, \Tcal) = \max_{\tau \in \mathcal{T}} R(\tau), \quad s.t. \quad C(\tau) = \kappa.$$
Note that their constraints and domains of $\kappa$ are different. 
All the functions characterize the shape of the dataset.
RF is ``local'' since it represents the highest reward of a cost and is only defined on reachable cost values in dataset $\mathcal{T}$. On the other hand, PF and IPF are ``global'', since PF/IPF is the supremum of all the RF values w.r.t costs smaller/larger than a cost threshold. They are both defined on a continuous space of $\kappa$.
% PF represents the reward upper bound of the trajectories that satisfy the constraint, while RF is the upper bound for the trajectories with a specific cost return.
It is also easy to observe that the Pareto frontier $\text{PF}(\kappa, \mathcal{T})$ is a non-deceasing function of $\kappa$, which suggests the trade-offs between safety and task performance: finding a policy with a small cost return usually needs to sacrifice the reward. Based on the definition of PF and IPF, we introduce $\epsilon$-reducible to characterize the property of the dataset.
\begin{definition}[$\epsilon$-reducible]
An offline safe RL dataset $\Tcal$ is $\epsilon$-reducible w.r.t. threshold $\kappa$ if: $\text{PF}(\kappa, \Tcal)=\text{IPF}(\kappa, \Tcal)+\epsilon$.
% i.e., $\exists \tau \in \Tcal, s.t. \ C(\tau) \geq \kappa; R(\tau) + \epsilon =\text{PF}(\kappa, \Tcal) $.
\end{definition}

It is worth noticing that $\epsilon \in \mathbb{R}$ rather than $\mathbb{R}_{\geq 0}$. 
A positive $\epsilon$ means that there does not exist any trajectory $\kappa \in \Tcal$ that can achieve a higher reward than $\text{PF}(\kappa, \Tcal)$ even if removing the safety constraint, so the optimal policy is more likely to be an interior point within the safety boundary. 
A negative $\epsilon$ indicates that the reward of most rewarding trajectories in $\Tcal$ is upper bounded by $\text{PF}(\kappa, \Tcal)-\epsilon$, and thus the agent has a high chance for violating safety constraint if the policy greedily maximizes the reward. In this case, the optimal policy will likely be on the safety constraint boundary.

\begin{figure}[ht]
\vspace*{-2mm}
\centering
\includegraphics[width=0.99\linewidth]{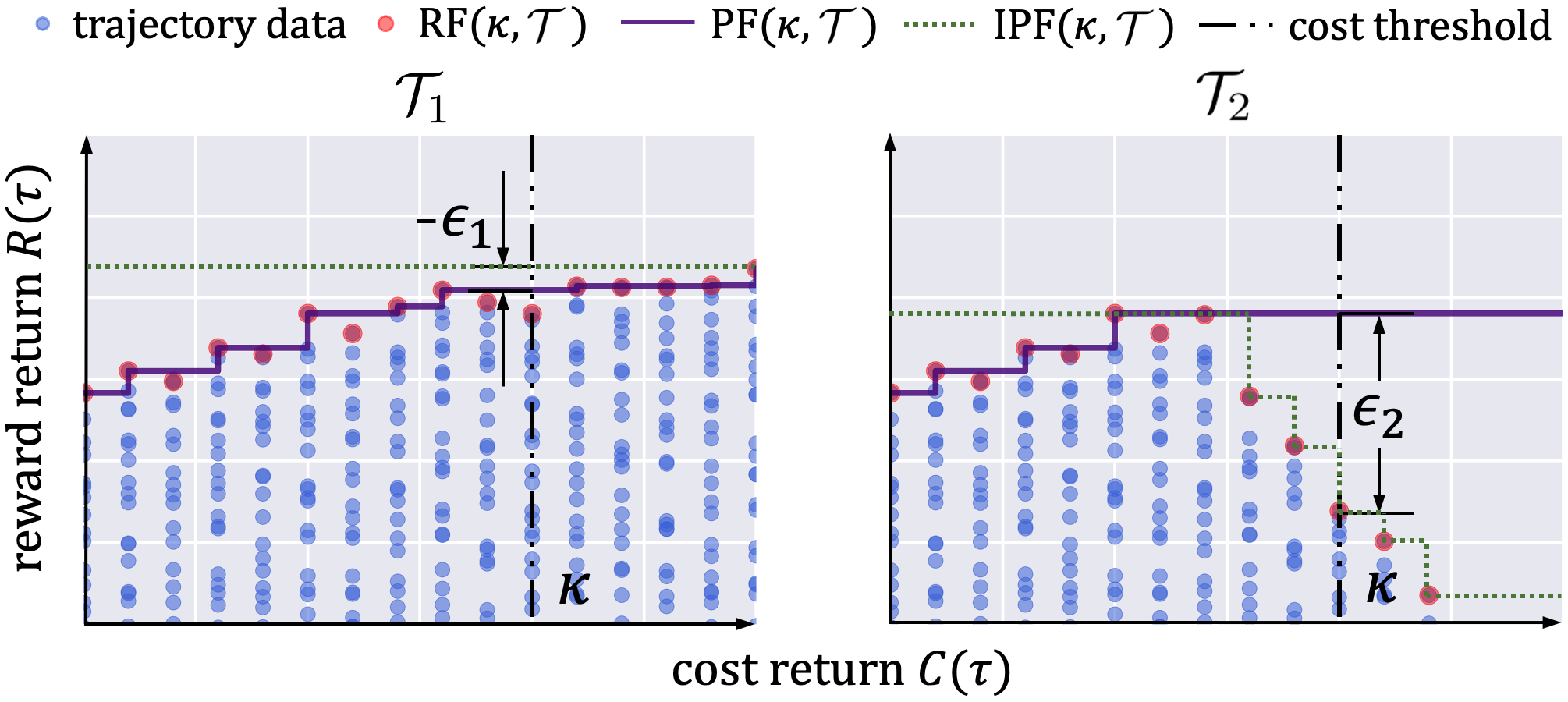}
\vspace*{-4mm}
\caption{Cost-reward return plot for two collected datasets $\Tcal_1, \Tcal_2$. Each point represents trajectories with corresponding episodic cost and reward values in the dataset.}
\label{fig:cr-plot-dataset}

\end{figure}

Fig.~\ref{fig:cr-plot-dataset} shows an example of the cost-reward return plots of two datasets $\mathcal{T}_1$ and $\mathcal{T}_2$.
Note that although $\Tcal_1$ and $\Tcal_2$ are collected in the same environment, $(\kappa, \Tcal_1)$ and $(\kappa, \Tcal_2)$ denote two different offline safe RL problems.
We observe that the $\epsilon$-reducible property can characterize the task difficulty.
For instance, problem $(\kappa, \Tcal_2)$ is usually easier to solve than $(\kappa, \Tcal_1)$, because $(\kappa, \Tcal_2)$ could be \textit{reduced} to an offline RL problem by simply maximizing the reward without considering the cost constraint.  
% We propose the $\epsilon$-reducible definition to characterize the problem difficulty for the same threshold $\kappa \in \mathbb{C}$ in the same environment $\Mcal$.
We have the following conjecture regarding the task difficulty:

% It is worth noticing that $\epsilon \in \mathbb{R}$ rather than $\mathbb{R}_{\geq 0}$. 
% A positive $\epsilon$ means that at least one safe trajectory exists, such that its reward is $\epsilon$ more than the reward frontier value at threshold $\kappa$: $\exists \tau \in \Tcal, s.t. \quad C(\tau) < \kappa; R(\tau)=\text{RF}(\kappa, \Tcal) + \epsilon, \epsilon>0$.
% It indicates that the most rewarding and safe trajectory is with a cost value less than $\kappa$, and thus the optimal policy is more likely to be an interior point within the safety boundary.
% On the other hand, a negative $\epsilon$ indicates the Pareto frontier value increases at trajectories with cost $\kappa$, which suggests that the optimal policy is likely to be on the constraint boundary.

\textit{Suppose problem $(\kappa, \Tcal)$ is $\epsilon$-reducible, then the smaller $\epsilon$, the more difficult to find the optimal solution.}

We empirically validate the hypothesis by experiments over different $\epsilon$-reducible problems, and interestingly, we find that \textbf{standard offline RL algorithms} can achieve safe performance with high-reward in large-$\epsilon$-reducible problems without solving constrained optimizations.
However, their performance deteriorates when the problem $(\kappa, \Tcal)$ possesses a smaller $\epsilon$ value. Detailed results and discussions can be found in Appendix \ref{app:data-reduction}.

\begin{remark}[Applicability]
    It is important to note that the comparison of $\epsilon$-reducible values is only valid within a single task under the same CMDP. Comparisons across different tasks are not meaningful as a smaller $\epsilon$ value in one task does not necessarily imply that this dataset is more challenging than another dataset related to a different task with a larger $\epsilon$.
\end{remark}

\begin{remark}[Limitations]
    The concept of $\epsilon$-reducibility may fall short when characterizing the complexity of noisy datasets or datasets related to highly-stochastic tasks. This is primarily because noisy datasets may include outlier trajectories with improbable high rewards and low costs. Similarly, in a highly-stochastic environment, the initial state distribution can significantly influence the final reward and cost. As such, the dataset is also likely to contain high-reward, low-cost trajectories due to ``lucky" initial conditions.
\end{remark}

\begin{remark}[Relation to Temptation]
    The concept of reducibility aligns with the temptation definition in safe RL literature~\cite{liu2022robustness}, as both describe the reward and cost trade-offs. However, they differ in their operational domains. Temptation focuses on the policy space and its expected returns, flagging a problem as tempting if it has high-reward but unsafe policies. In contrast, $\epsilon$-reducibility evaluates the dataset in the trajectory space, labeling a dataset as small $\epsilon$-reducible if it includes high-reward, high-cost trajectories. Hence, they offer complementary perspectives to understand the challenges of safe RL.
\end{remark}

The proposed $\epsilon$-reducible can serve as a measure of the offline safe RL problem difficulties: larger $\epsilon$ means the problem is more likely to be \textbf{reduced} to standard offline RL, though more rigorous analysis remains to be explored in future work.
% The reducible property is related to the temptation definition in the safe RL literature~\cite{liu2022robustness}, which describes the fact that the policy might be tempted during training to achieve high rewards but violate constraints. 
In this work, we are more interested in small $\epsilon$-reducible problems, because these problems can hardly be solved by standard offline RL methods.  
The cost-reward return plots and their RF curves of the datasets in our experiments are also presented in Appendix~\ref{section: RF curves}.

\subsection{Offline Safe RL Beyond a Single Threshold}
\label{sec:osrl-beyond-one-thres}

Most existing offline safe RL approaches train policies by solving a constrained optimization problem, where learnable dual variables are updated based on the estimation of constraint violation cost and a target threshold~\cite{xu2022constraints, lee2022coptidice, polosky2022constrained}.
The cost return estimation is of the form $C_\pi = \E_{\tau\sim\pi}[C(\tau)]$ and the reward return is $R_\pi = \E_{\tau\sim\pi}[R(\tau)]$. 
The learning algorithm aims to train a policy $\pi$ to maximize the reward return $R_\pi$ while satisfying the constraint $C_\pi \leq \kappa$.
The constrained optimization training scheme works well in online safe RL settings~\cite{stooke2020responsive}, however, two main challenges arise for the offline setting. We detail them as follows.

First, \textbf{the trained policy tends to be either unsafe or overly conservative} due to biased and inaccurate estimation in the offline setting.
This is because the trajectories $\tau$ from the dataset $\Tcal$ are sampled from various unknown behavior policies rather than the optimized policy.
In RL, biased estimation of the reward return $R_\pi$ may not affect the results because the maximization operation over $R_\pi$ is invariant to the bias and scale,
However, in safe RL, a biased estimation of cost $C_\pi$ could cause significantly wrong dual variables, because its absolute value is compared against a fixed threshold $\kappa$.
A small negative bias can lead to unsafe behaviors, and a positive bias can induce a conservative policy.
This problem is challenging for off-policy safe RL~\cite{liu2022constrained} and is more difficult to address in the offline setting, as we will show empirically in the experiment section~\ref{sec:exp1}.

Second, \textbf{the trained policy cannot be easily adapted to different constraint thresholds without re-training}. 
The cost threshold needs to be pre-selected and kept fixed throughout training because otherwise, the dual variables for penalizing constraint violations could be unstable when solving the constrained optimization and thus diverge the learning process.
Therefore, adapting the policy to different constraint conditions requires re-training with new thresholds.

The second challenge corresponds to the problem of learning a safe policy from the offline dataset beyond a single constraint threshold.
Formally speaking, given a dataset $\Tcal$, the trained agent $\pi(a|s,\kappa)$ is expected to be generalized to arbitrary cost thresholds $\forall \kappa \in [C_{\text{min}}, C_{\text{max}}]$, where $C_{\text{min}}, C_{\text{max}}$ are the minimum and maximum of the cost return of the trajectories in the dataset.
The best reward return of $\pi(a|s,\kappa)$ is lower-bounded by the Pareto frontier value of the dataset with threshold $\kappa$: $\text{PF}(\kappa, \mathcal{T})$.

The limitations of the constrained optimization-based training paradigm motivate us to think about other learning schemes.
We find that sequential modeling techniques, such as Decision Transformer (DT)~\cite{chen2021decision}, have great potential to achieve zero-shot adaptation to different constraint thresholds while maintaining safety and near Pareto optimal task performance, 
As introduced in Sec.~\ref{sec:dt}, the DT policy predicts target-return-conditioned actions, which provides the flexibility to adjust the agent behaviors after training.
However, we observe that simply adding a target cost return token to the input sequence of DT can hardly ensure safety in practice.
Therefore, we propose two simple yet effective improvements over DT to train a safe and adaptive policy, which yields the Constrained Decision Transformer (CDT) algorithm.

\subsection{Constrained Decision Transformer}
\label{sec:cdt}

\begin{figure}[ht]
\vspace*{-1mm}
\centering
\includegraphics[width=0.99\linewidth]{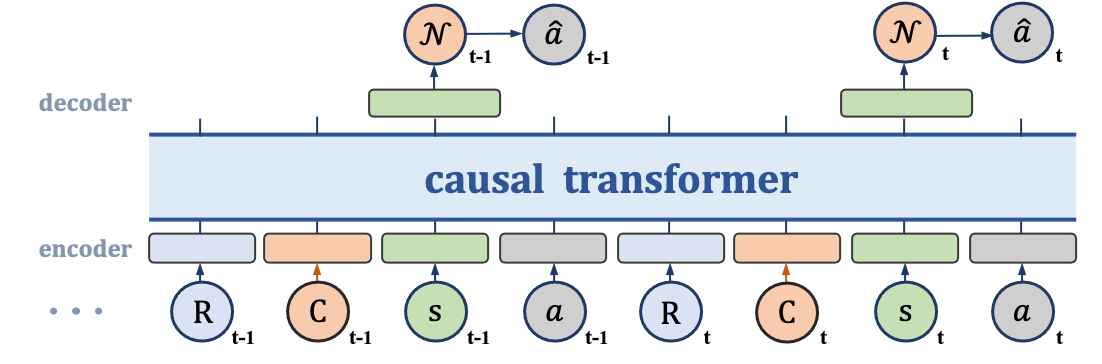}
\vspace*{-4mm}
\caption{Constrained decision transformer architecture.}
\label{fig:cdt}
\end{figure}

CDT is built upon the DT architecture with two main different components:
1) stochastic policy with entropy regularization, and 2) Pareto frontier-oriented data augmentation by target return relabeling.
We found that the two techniques play crucial roles in improving safety and robustness against conflicting target returns.
We detail them as follows.

\textbf{Stochastic CDT policy with entropy regularization.}
The model architecture of CDT is shown in Fig.~\ref{fig:cdt}, where the differences between CDT and DT are highlighted in orange.
Recall that the input tokens for DT are 
$\mathbf{R}_{-K:t}=\{R_K, ..., R_t\}$, $\mathbf{s}_{-K:t}$ and $\mathbf{a}_{-K:t-1}$, where $K\in\{1, ..., t-1\}$ is the context length and $R_t=\sum_{t'=t}^T r_{t'}$ is the reward return of step $t$.
CDT augments the input sequences with an additional element that represents the target cost threshold $\mathbf{C}_{-K:t}=\{C_K, ..., C_t\}$, where $C_t=\sum_{t'=t}^T c_{t'}$ is the cost return starting from timestep $t$.
The intuition is to generate actions conditioned on both the reward return and the cost return.
For example, at timestep $t$, setting $C_t=10$ and $R_t=80$ means that we expected the agent to obtain $80$ rewards with a maximum allowed $10$ costs.
Different from DT, which predicts deterministic action sequences, CDT adopts the stochastic Gaussian policy representation, drawing inspiration from the Online Decision Transformer architecture \cite{zheng2022online}.
Denote $\mathbf{o}_t :=\{ \mathbf{R}_{-K:t}, \mathbf{C}_{-K:t}, \mathbf{s}_{-K:t}, \mathbf{a}_{-K:t-1}\}$ as the input tokens and $\theta$ as the CDT policy parameters, we have:
$$\pi_\theta (\cdot | \mathbf{o}_t) = \mathcal{N}\left( \mu_{\theta}(\mathbf{o}_t), \Sigma_\theta(\mathbf{o}_t) \right).$$

The use of a stochastic representation confers multiple benefits. Firstly, a deterministic policy may be more prone to producing out-of-distribution actions due to systematic bias, which can lead to large compounding errors in an offline environment and potentially result in constraint violations~\cite{xu2022constraints}. Further details on this property can be found in Appendix \ref{app:stochastic-policy}. 
Secondly, the stochastic policy representation allows the policy to explore a more diverse range of actions and enhance performance through interaction with the environment. This aligns with the pretraining and fine-tuning learning paradigm in the literature and shows great promise for real-world applications \cite{zheng2022online}.
Finally, we can easily apply a regularizer to prevent the policy from overfitting and improve the robustness against approximation errors \cite{ziebart2010modeling, eysenbach2021maximum}. 
We adopt the Shannon entropy regularizer $H[\pi_\theta(\cdot | \mathbf{o})]$ for CDT, which is widely used in RL~\cite{haarnoja2018soft}. 
The optimization objective is to minimize the negative log-likelihood loss while maximizing the entropy with weight $\lambda \in [0, \infty)$:
\begin{equation}
 \begin{split}
    \min_\theta \quad \mathbb{E}_{\mathbf{o}\sim \mathcal{T}}  \left[ -\log \pi_\theta (\mathbf{a} | \mathbf{o}) - \lambda H[\pi_\theta(\cdot | \mathbf{o})] \right]
 \end{split}
 \label{eq:cdt-loss}
\end{equation}
Since CDT adopts the target returns-conditioned policy structure, the agent behavior is sensitive to the choices of target reward and cost.
In offline RL, one can set a large enough target reward for the agent to maximize the reward.
However, as shown in Fig.~\ref{fig:cr-plot-dataset}, the feasible choices of valid target cost and reward return pairs are restricted under the RF points, which brings a major challenge for CDT: how can we resolve the potential conflict between desired returns and ensure the target cost is of higher priority than the target reward?
For instance, if the initial reward return is set to be slightly higher than the RF value of the initial target cost threshold, then it is hard to determine whether the policy will achieve the desired reward but violate the constraint or satisfy the cost threshold but with a lower reward.

\begin{algorithm}[h]
\caption{Data Augmentation via Relabeling}
{\bfseries Input:} \raggedright dataset $\Tcal$, samples $N$, reward sample max $r_{max}$ \par
{\bfseries Output:} \raggedright augmented trajectory dataset $\Tcal$ \par
\begin{algorithmic}[1] % The number tells where the line numbering should start
% \STATE $\triangleright$ \textit{compute cost sampling range}
\STATE $c_{min} \leftarrow \min_{\tau\sim\Tcal} C(\tau)$, $c_{max} \leftarrow \max_{\tau\sim\Tcal} C(\tau)$
% \STATE Sample $N$ infeasible returns $\{(\rho_i, \kappa_i)\}_{i=1}^N$
\FOR{$i=1,..., N$}
\STATE $\triangleright$ \textit{sample a cost return}
\STATE $\kappa_i \sim \text{Uniform}(c_{min}, c_{max})$
\STATE $\triangleright$ \textit{sample a reward return above the RF value}
\STATE $\rho_i \sim \text{Uniform}(\text{RF}(\kappa_i, \Tcal), r_{max})$
\STATE $\triangleright$ \textit{find the closest and safe Pareto trajectory}
\STATE $\tau^*_i \leftarrow \arg\max_{\tau\sim\Tcal} R(\tau), s.t. \quad C(\tau) \leq \kappa_i$
\STATE $\triangleright$ \textit{relabel the reward and cost return}
\STATE $\hat{\tau}_i \leftarrow \{\mathbf{R}^*_i + \rho_i - R(\tau^*_i),\mathbf{C}^*_i+ \kappa_i - C(\tau^*_i),\mathbf{s}^*_i,\mathbf{a}^*_i\}$
\STATE $\triangleright$ \textit{append the trajectory to the dataset}
\STATE $\Tcal \leftarrow \Tcal \cup \{\hat{\tau}_i\}$
\ENDFOR
\end{algorithmic} \label{algo:augmentation}
\end{algorithm}

\begin{figure}[h]
\vspace*{-1mm}
\centering
\includegraphics[width=0.99\linewidth]{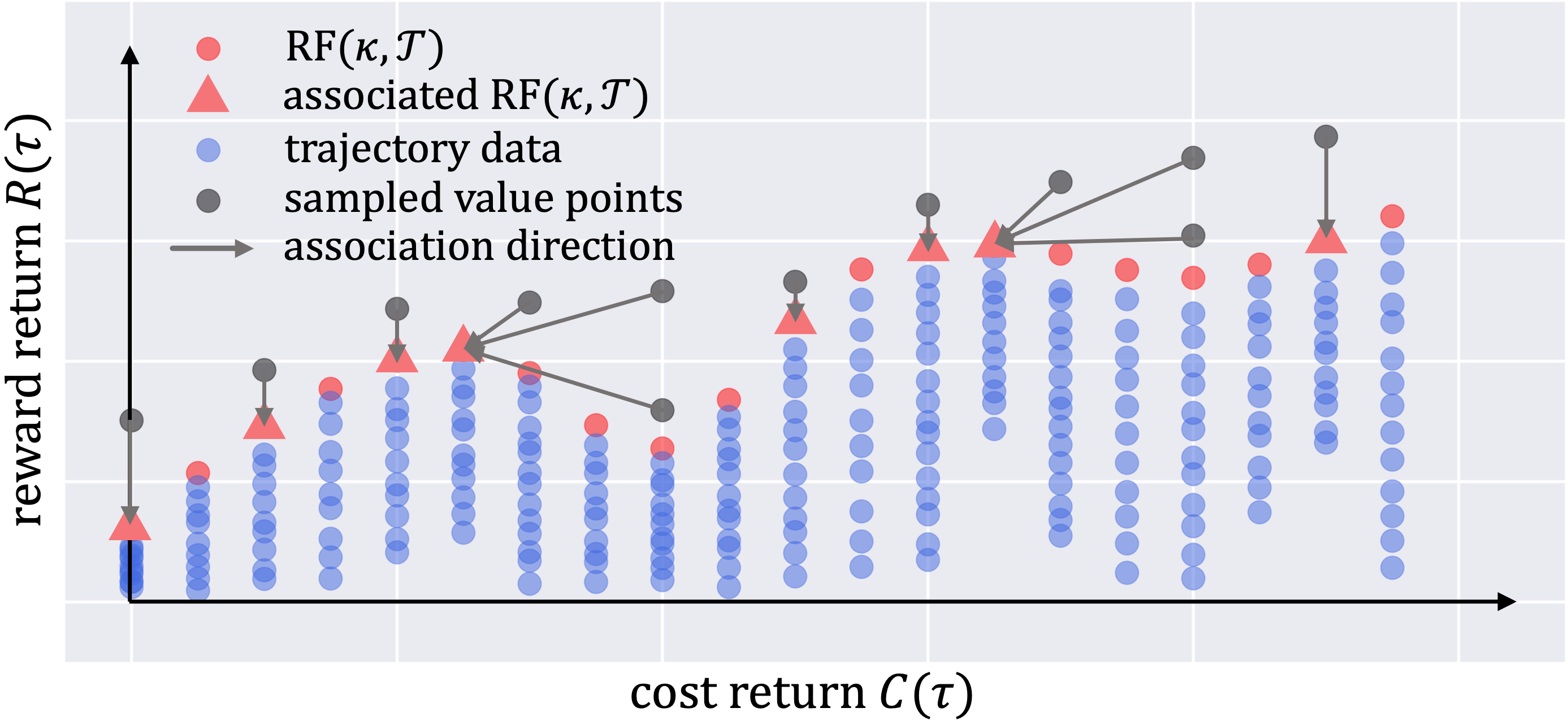}
\vspace*{-2mm}
\caption{Data augmentation.}
\label{fig:augmentation}
\vspace*{-1mm}
\end{figure}

\textbf{Data augmentation by return relabeling.}
We propose an effective augmentation technique to address the above issue by utilizing the Pareto frontier and reward frontier properties of the trajectory-level dataset $\Tcal$.
Suppose $(\rho, \kappa)$ is an infeasible target return pair, i.e., $\rho > \text{RF}(\kappa, \Tcal)$.
We associate the conflict target with the safe trajectory that of the maximum reward return: $\tau^* = \arg\max_{\tau\sim\Tcal} R(\tau), s.t. \quad C(\tau) \leq \kappa$.
We can observe that $\tau^* = \{\mathbf{R}^*,\mathbf{C}^*,\mathbf{s}^*,\mathbf{a}^*\}$ is the maximum-reward Pareto optimal trajectory with cost less than $\kappa$.
Then we append the new trajectory data $\hat{\tau} = \{\mathbf{R}^* + \rho - R(\tau^*),\mathbf{C}^*+ \kappa - C(\tau^*),\mathbf{s}^*,\mathbf{a}^*\}$ to the dataset: $\Tcal \leftarrow \Tcal \cup \{\hat{\tau}\}$.
Note that the operators over $\mathbf{R}^*$ and $\mathbf{C}^*$ are element-wise.
%where $R(\tau^*) = R_1^* \in \mathbf{R}^*$ and $C(\tau^*) = C_1^* \in \mathbf{C}^*$.
The intuition is to relabel the associated Pareto trajectory's reward and cost returns, such that the agent can learn to imitate the behavior of the most rewarding and safe trajectory $\tau^*$ when the desired return $(\rho,\kappa)$ is infeasible.
Fig.~\ref{fig:augmentation} shows an example of the procedure, where the arrows associate Pareto-optimal trajectories with corresponding augmented return pairs.
The detailed augmentation procedures are presented in Algorithm~\ref{algo:augmentation}.
% There are three steps: 1) random sample infeasible reward-cost return pairs that are above the reward frontier. 2) associate each sampled return with the closest Pareto-optimal trajectory that satisfies constraints. 3) relabel the returns and append the trajectories to the dataset. 

It is worth noting that real-world datasets can be noisy, occasionally including anomalous "lucky" trajectories that record high reward and low-cost returns despite originating from subpar behavioral policies. These outliers, while rare, can disrupt the data augmentation procedure, thereby negatively affecting CDT's performance. To address this issue, our implementation utilizes two specific techniques. The first involves associating each augmented return pair $(r, c)$ with a trajectory sampled in proximity to the nearest and safe Pareto frontier data point, based on a specified distance metric. The second technique employs a density filter to remove such outliers exhibiting abnormal reward and cost returns during the creation of the training dataset, thus mitigating the outlier concern. 
More details regarding these two techniques and empirical validations are available in Appendix \ref{app:noisy-dataset}.

\textbf{Training and evaluation.} CDT generally follows the training and evaluation schemes of return-conditioned sequential modeling methods~\cite{chen2021decision, zheng2022online}.
The training procedure is similar to training a Transformer in supervised learning: sample a batch of sequences $\mathbf{o}, \mathbf{a}$ from the augmented dataset $\Tcal$, compute the loss in Eq. (\ref{eq:cdt-loss}) to optimize the Transformer policy model $\pi_\theta$ via gradient descent.
The evaluation procedure for a trained CDT model is presented in Algorithm~\ref{algo:evaluation}.
Note that it differs from standard RL, where the policy directly predicts the action based on the state.
As shown in Fig.~\ref{fig:cdt}, the input for the return-conditioned policy is a tuple of four sequences: target reward and cost returns for each step, past states, and actions.
Therefore, the output is also a sequence of actions, but we only execute the last one in the environment. The target returns will be updated correspondingly upon receiving new reward and cost signals from the environment.

\begin{algorithm}[h]
\caption{Returns Conditioned Evaluation for CDT}
{\bfseries Input:} \raggedright trained Transformer policy $\pi_\theta$, episode length $T$, context length $K$, target reward and cost $R_1, C_1$, env \par
\begin{algorithmic}[1]
\STATE Get the initial state: $s_1 \leftarrow \text{env.reset}()$
\STATE Initialize input sequence $\mathbf{o}=[\{R_1, C_1, s_1\}]$
\FOR{$t=1,..., T$}
\STATE Get predicted action $a_t \sim \pi(\cdot | \mathbf{o}[-K:])[-1]$
\STATE Execute the action: $s_{t+1}, r_t, c_t \leftarrow \text{env.step}(a_t)$
\STATE $\triangleright$ \textit{compute target returns for the next step}
\STATE $R_{t+1} = R_t-r_t, C_{t+1} = C_t-c_t,$
\STATE Append the new token $o_t = \{R_{t+1},C_{t+1}, s_{t+1}, a_t \}$ to the sequence $\mathbf{o}$
\ENDFOR
\end{algorithmic} \label{algo:evaluation}
\end{algorithm}

\vspace{-1mm}
\section{Experiment}
\vspace{-1mm}
\label{sec:experiment}
In this section, we aim to evaluate the proposed approach and empirically answer the following questions: 1) can CDT learn a safe policy from a small $\epsilon$ reducible offline dataset? 2) what is the importance of each component in CDT? 3) can CDT achieve zero-shot adaption to different constraint thresholds? 4) is CDT robust to conflict reward returns? To address these questions, we adopt the following tasks to evaluate CDT and baseline approaches.

\begin{table*}[ht]
\centering
\renewcommand{\arraystretch}{1.1}
\resizebox{1.\linewidth}{!}{
% \begin{tabular}{c|cc|cc|cc|cc|cc}
\begin{tabular}{c|cc|cc|cc|cc|cc|cc}
% \hline
\toprule
                          & \multicolumn{2}{c|}{Ant-Run}                                                  & \multicolumn{2}{c|}{Car-Circle}                                               & \multicolumn{2}{c|}{Car-Run}                                                  & \multicolumn{2}{c|}{Drone-Circle}                                             & \multicolumn{2}{c|}{Drone-Run}                                                & \multicolumn{2}{c}{Average}                                                  \\
\multirow{-2}{*}{Methods} & reward $\uparrow$                                & cost $\downarrow$                                 & reward $\uparrow$                                & cost $\downarrow$                                 & reward $\uparrow$                                & cost $\downarrow$                                 & reward $\uparrow$                                & cost $\downarrow$                                 & reward $\uparrow$                                & cost $\downarrow$                                 & \multicolumn{1}{c}{reward $\uparrow$}               & \multicolumn{1}{c}{cost $\downarrow$}             \\ \hline
CDT(ours)                 & {\color[HTML]{0000FF} \textbf{89.76}} & {\color[HTML]{0000FF} \textbf{0.83}} & {\color[HTML]{0000FF} \textbf{89.53}} & {\color[HTML]{0000FF} \textbf{0.85}} & \textbf{99.0}                         & \textbf{0.45}                        & {\textbf{73.01}} & {\textbf{0.88}} & {\color[HTML]{0000FF} \textbf{63.64}} & {\color[HTML]{0000FF} \textbf{0.58}} & {\color[HTML]{0000FF} \textbf{82.99}} & {\color[HTML]{0000FF} \textbf{0.72}} \\
BC-Safe                   & \textbf{80.56}                        & \textbf{0.64}                        & \textbf{78.21}                        & \textbf{0.74}                        & \textbf{97.21}                        & \textbf{0.01}                        & \textbf{66.49}                        & \textbf{0.56}                        & \textbf{32.73}                        & \textbf{0.0}                         & \textbf{71.04}                        & \textbf{0.39}                        \\
DT-Cost                   & {\color[HTML]{656565} 91.69}          & {\color[HTML]{656565} 1.32}          & {\color[HTML]{656565} 89.08}          & {\color[HTML]{656565} 2.14}          & {\color[HTML]{656565} 100.67}         & {\color[HTML]{656565} 11.83}         & {\color[HTML]{656565} 78.09}          & {\color[HTML]{656565} 2.38}          & {\color[HTML]{656565} 72.3}           & {\color[HTML]{656565} 4.43}          & {\color[HTML]{656565} 86.37}          & {\color[HTML]{656565} 4.42}          \\
BCQ-Lag                   & {\color[HTML]{656565} 92.7}           & {\color[HTML]{656565} 1.04}          & {\color[HTML]{656565} 89.76}          & {\color[HTML]{656565} 3.91}          & {\color[HTML]{656565} 96.14}          & {\color[HTML]{656565} 3.21}          & {\color[HTML]{656565} 71.14}          & {\color[HTML]{656565} 3.37}          & {\color[HTML]{656565} 47.61}          & {\color[HTML]{656565} 1.81}          & {\color[HTML]{656565} 79.47}          & {\color[HTML]{656565} 2.67}          \\
BEAR-Lag                  & {\color[HTML]{656565} 91.19}          & {\color[HTML]{656565} 1.66}          & {\color[HTML]{656565} 15.48}          & {\color[HTML]{656565} 2.24}          & {\color[HTML]{0000FF} \textbf{99.09}} & {\color[HTML]{0000FF} \textbf{0.09}} & {\color[HTML]{656565} 72.36}          & {\color[HTML]{656565} 1.99}          & \textbf{19.07}                        & \textbf{0.0}                         & {\color[HTML]{656565} 59.44}          & {\color[HTML]{656565} 1.2}           \\
CPQ                       & \textbf{78.52}                        & \textbf{0.14}                        & \textbf{75.99}                        & \textbf{0.0}                         & \textbf{97.72}                        & \textbf{0.11}                        & {\color[HTML]{656565} 55.14}          & {\color[HTML]{656565} 9.67}          & {\color[HTML]{656565} 72.24}          & {\color[HTML]{656565} 4.28}          & {\color[HTML]{656565} 75.92}          & {\color[HTML]{656565} 2.84}          \\ 
COptiDICE                 & \textbf{45.55}                        & \textbf{0.6}                         & {\color[HTML]{656565} 52.17}          & {\color[HTML]{656565} 6.38}          & \textbf{92.86}                        & \textbf{0.89}                        & {\color[HTML]{656565} 36.44}          & {\color[HTML]{656565} 5.54}          & {\color[HTML]{656565} 26.56}          & {\color[HTML]{656565} 1.38}          & {\color[HTML]{656565} 50.72}          & {\color[HTML]{656565} 2.96}          \\ \hline
CDT(w/o augment)          & {\color[HTML]{656565} 93.62}          & {\color[HTML]{656565} 1.53}          & {\color[HTML]{656565} 89.8}           & {\color[HTML]{656565} 1.38}          & {\color[HTML]{656565} 99.58}          & {\color[HTML]{656565} 1.89}          & {\color[HTML]{656565} 74.9}           & {\color[HTML]{656565} 1.35}          & {\color[HTML]{656565} 66.93}          & {\color[HTML]{656565} 1.53}          & {\color[HTML]{656565} 84.97}          & {\color[HTML]{656565} 1.54}          \\
CDT(w/o entropy)          & \textbf{87.47}                        & \textbf{0.64}                        & {\color[HTML]{656565} 89.94}          & {\color[HTML]{656565} 1.07}          & \textbf{98.92}                        & \textbf{0.44}                        & {\textbf{73.76}} & {\textbf{0.97}}    & \textbf{62.29}                       & \textbf{0.6}                          & \textbf{82.48}                       & \textbf{0.74}                        \\
CDT(deterministic)        & {\color[HTML]{656565} 94.21}          & {\color[HTML]{656565} 1.42}          & {\color[HTML]{656565} 89.53}          & {\color[HTML]{656565} 1.43}          & {\color[HTML]{656565} 101.52}         & {\color[HTML]{656565} 17.53}         & {\color[HTML]{0000FF} \textbf{76.4}}  & {\color[HTML]{0000FF} \textbf{1.0}}  & {\color[HTML]{656565} 68.44}          & {\color[HTML]{656565} 1.36}          & {\color[HTML]{656565} 86.02}          & {\color[HTML]{656565} 4.55}          \\
\bottomrule % \hline
\end{tabular}
}
\caption{Evaluation results of the normalized reward and cost. The cost threshold is 1.
$\uparrow$: the higher reward, the better. $\downarrow$: the lower cost (up to the threshold 1), the better. 
Each value is averaged over 20 episodes and 3 seeds.
\textbf{Bold}: Safe agents whose normalized cost is smaller than 1. 
{\color[HTML]{656565} Gray}: Unsafe agents.
{\color[HTML]{0000FF} \textbf{Blue}}: Safe agent with the highest reward.
}
\label{tab:exp-results}
\end{table*}

\textbf{Tasks.} 
We use several robot locomotion continuous control tasks that are commonly used in previous works \citep{achiam2017constrained, chow2019lyapunov, zhang2020first}. 
The simulation environments are from a public benchmark \cite{gronauer2022bullet}. 
We consider two environments (\texttt{Run} and \texttt{Circle}) and train multiple different robots (\texttt{Car, Drone}, and \texttt{Ant}). 
In the Run environment, the agents are rewarded for running fast between two boundaries and are given constraint violation cost if they run across the boundaries or exceed an agent-specific velocity threshold. 
In the Circle environment, the agents are rewarded for running in a circle but are constrained within a safe region that is smaller than the radius of the target circle. 
We name the task as \texttt{robot-environment} such as \texttt{Ant-Run}.

\textbf{Offline datasets.}
The dataset format follows the D4RL benchmark \cite{fu2020d4rl}, where we add another cost entry to record binary constraint violation signals.
We collect offline datasets using the CPPO safe RL approach with well-tuned hyperparameters~\cite{stooke2020responsive}.
We gradually increase its cost threshold such that the trajectories can cover a diverse range of cost returns and reward returns.
All the training data for CPPO is stored as the raw dataset, which may contain many repeated trajectories.
We further down-sample the data by applying a grid filter over the cost-reward return space (Fig.~\ref{fig:cr-plot-dataset}) and trim redundant trajectories to avoid the impact of unevenly distributed data~\cite{gulcehre2020rl, gong2022mind, singh2022offline}.
Namely, we divide the cost-reward space into multiple 2D grids, randomly select a fixed number of trajectories within each grid and discard the remaining ones.
The cost-return plots of different datasets used in this work are presented in Appendix~\ref{app:more-results}.
% We will also release the datasets and open-source the code as a public benchmark for the offline safe RL problem.

% Motivated by the D4RL benchmark \cite{fu2020d4rl}, we first use the PPO-Lagrangian (PPO-Lag) \cite{stooke2020responsive} to train the agents with an increasing cost threshold that is proportional to the training steps and record all the training data as the raw data. 
% Considering the quality of the raw data, such as the unevenly distributed trajectories in reward and cost return space could have an adverse impact on offline training \cite{gulcehre2020rl, gong2022mind, singh2022offline}, 
% we down-sample the raw data by dividing it evenly into grids according to episode reward and cost return and randomly select a fixed number of trajectories within each grid.
% Lastly, we select the trajectories whose cost return lies within a given range as the offline datasets.

\textbf{Metrics.} We adopt the normalized reward return and the normalized cost return as the comparison metrics, which are consistent with the offline RL literature~\cite{fu2020d4rl}.
Denote $r_{\text{max}}(\Tcal)$ and $r_{\text{min}}(\Tcal)$ as the maximum reward return and the minimum reward return in dataset $\Tcal$. The normalized reward is computed by:
$$
R_{\text{normalized}} = \frac{R_\pi - r_{\text{min}}(\Tcal)}{r_{\text{max}}(\Tcal)-r_{\text{min}}(\Tcal)} \times 100,
$$
where $R_\pi$ denotes the evaluated reward return of policy $\pi$.
The normalized cost is defined a bit differently from the reward, which is computed by the ratio between the evaluated cost return $C_\pi$ and the target threshold $\kappa$:
$$
C_{\text{normalized}} = \frac{C_\pi}{\kappa + \epsilon},
$$
where $\epsilon$ is a small positive number to ensure numerical stability if the threshold $\kappa=0$. Note that the cost return is always non-negative in our setting, and we use $\kappa=10$ by default.
Without otherwise statements, we will abbreviate ``normalized cost return" as ``cost" and ``normalized reward return" as ``reward" for simplicity.

We can observe that a policy is unsafe if the cost is greater than $1$.
We deliberately scale the reward around the range $[0, 100]$ and the cost around $1$ to distinguish them in the result table better.
The comparison criteria follow the safe RL setting~\cite{ray2019benchmarking}: a safe policy is better than an unsafe one.
For two unsafe policies, the one with a lower cost is better.
For two safe policies, the one with a higher reward is better.
% To be consistent with the standard offline RL, we compare the methods in terms of the normalized reward and cost return by dividing the real reward and cost return of a trajectory by the maximum reward return in the corresponding data and the cost threshold during testing respectively. 
% We also scale the normalized reward return by multiplying 100 to make the results more clear.
% We consider an agent safe if its normalized cost return is smaller than 1 and an agent has the best performance if it is both safe and achieves the highest normalized reward return.

\textbf{Baselines with a fixed cost threshold.}
We use two recent offline safe RL approaches:  \textbf{CPQ} \cite{xu2022constraints} and \textbf{COptiDICE} \cite{lee2022coptidice} as two strong baselines.
We adopt two Lagrangian-based baselines: BCQ-Lagrangian (\textbf{BCQ-Lag}) and BEAR-Lagrangian (\textbf{BEAR-Lag}), which is built upon BCQ \cite{fujimoto2019off} and BEAR \cite{kumar2019stabilizing}, respectively.
The Lagrangian approach follows the expert policy CPPO implementation, which uses adaptive PID-based Lagrangian multipliers to penalize constraint violations \cite{stooke2020responsive}.
We use the vanilla Decision Transformer~\cite{chen2021decision} with an additional cost return token as another baseline \textbf{DT-Cost}, aiming to compare the effectiveness of the proposed CDT training techniques. 
We also include a Behavior Cloning baseline (\textbf{BC-Safe}) that only uses safe trajectories to train the policy. 
This serves to measure whether each method actually performs effective RL, or simply copies the data.

We also conducted comprehensive studies on the Behavior Cloning method with different datasets, including \textbf{BC-all, BC-risky, BC-frontier,} and \textbf{BC-boundary}. Due to space constraints, we defer the visualization of datasets and experiment results on these BC-variants to Appendix \ref{app:more-bc}.
More comprehensive details of dataset collection and baseline implementations are available at \cite{liu2023datasets}.

\textbf{Hyperparameters.} We use a fixed set of hyperparameters for CDT across all tasks. Most common parameters, such as the gradient steps, are also the same for CDT and baselines. The detailed hyperparameters are in Appendix \ref{app:exp-hyperparameters}.

\begin{figure*}[!ht]
\centering
\includegraphics[width=0.95\textwidth]{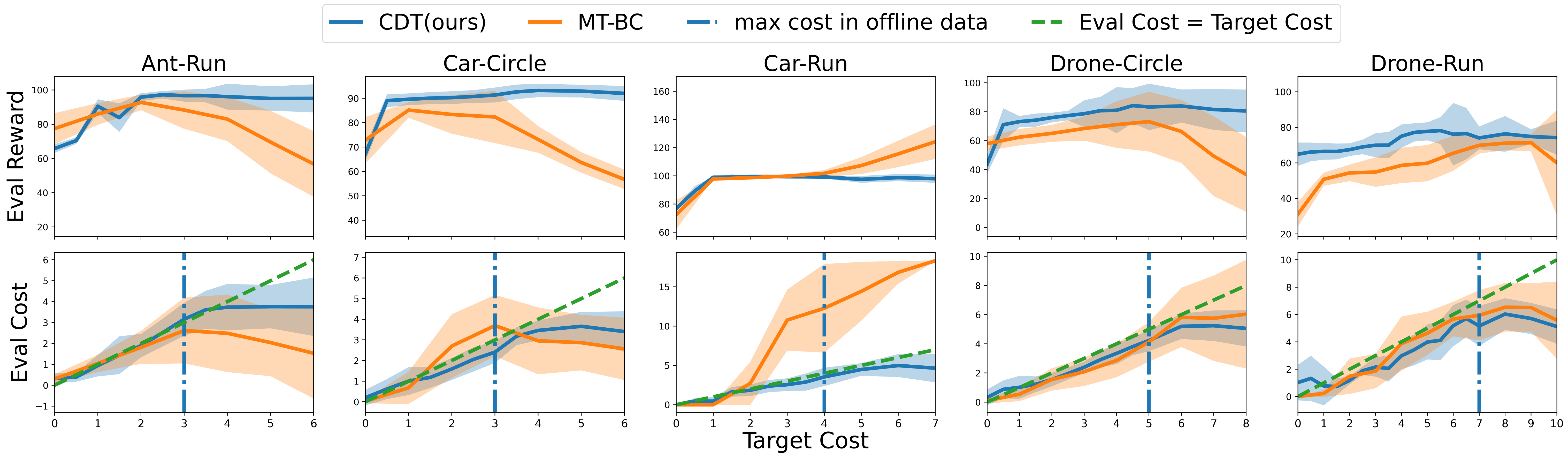}
\vspace{-4mm}
\caption{Results of zero-shot adaption to different cost returns. Each column is a task. 
The x-axis is the target cost return.
The first row shows the evaluated reward, and the second row shows the evaluated cost under different target costs. 
All plots are averaged among 3 random seeds and 20 trajectories for each seed. 
The solid line is the mean value, and the light shade represents the area within one standard deviation.
% The green dashed line . 
The vertical line is the maximum normalized cost return in the offline dataset.}
\label{fig:zero-shot}
\vspace{-1mm}
\end{figure*}

\subsection{Can CDT learn safe policies from offline datasets?}
\label{sec:exp1}

The evaluation results for different trained policies are presented in Table \ref{tab:exp-results}. 
We can find that only our method (CDT) and BC-Safe successfully learn safe policies for all tasks, while CDT consistently achieves higher reward returns than BC-safe.
It makes sense since BC-safe only uses safe data to train and thus fails to explore high-rewarding trajectories.
The comparison between BC-safe indicates that CDT performs effective RL rather than copying data.

The results of DT-cost show that simply adding a cost return token to the original DT structure can not train a constraint satisfaction policy, though it successfully learns to maximize the reward return. Note that in the Car-Run task, DT-cost even outperforms the best trajectory's reward in the dataset; however, the cost is also extremely high.
The comparison indicates that the proposed training techniques in CDT are crucial in learning a safe policy.

The Lagrangian-based baselines BCQ-Lag and BEAR-Lag fail to behave safely on most tasks, which suggests that directly applying widely-used safe RL techniques to the offline setting can hardly work well.
Surprisingly, the CPQ and COptiDICE methods that are designed for offline safe RL also fail to satisfy the constraints in difficult Drone-related tasks.
Particularly, the CPQ algorithm either performs over-conservatively with near zero cost or too aggressively with large costs.
As we discussed in Sec.~\ref{sec:osrl-beyond-one-thres}, one reason for the poor performance is the difficult-to-estimate cost value of the optimized policy. 
The trajectories in the dataset are collected from various behavior policies and may cause biased cost value estimation, which then causes too large or too small dual variables.
Accurately fitting a cost critic is still a challenging problem for off-policy safe RL~\cite{liu2022constrained}, let alone the offline setting.
The poor performance of baselines shows the difficulties of the experimental tasks; however, the proposed CDT can learn safe and high-rewarding policies in those tasks very well.
% The poor safety performance of DT-Cost shows that simply adding the target cost return does not work well in the offline safe RL setting.
% We can observe that BCQ-Lag fails to learn a safe policy in all the tasks.
% The BEAR-Lag, CPQ and COptiDICE struggle to meet the safety constraint and have difficulity learning high-rewarding policies due to the biased and inaccurate estimation of the reward and cost Q values in the offline setting.
% By comparing BC-Safe and CDT, we can observe the trade-offs between task and safety performance. 
% Although BC-Safe achieves safe policies and has a lower normalized cost return, it sacrifices reward for safety and thus is too conservative and has a lower reward. 
% On the other hand, CDT is able to learn constraint satisfaction policy from the offline dataset and obtain higher rewards in most tasks and achieves the highest reward among the safe agents in average.
% As a result, CDT outperforms all the baselines in terms of both task and safety performance.

\textbf{Ablation study.}
To study the influence of data augmentation, stochastic policy, and entropy regularization, we conduct experiments by removing each component from CDT.
The results are shown in the lower part of Table \ref{tab:exp-results}.
It is clear to see both augmentation and stochastic representation are necessary and important components since we can observe significant safety performance degradation if removing either one of them. 
Besides, entropy regularization can result in a slight improvement regarding the overall performance.

\begin{figure*}[h]
\centering
\includegraphics[width=0.95\linewidth]{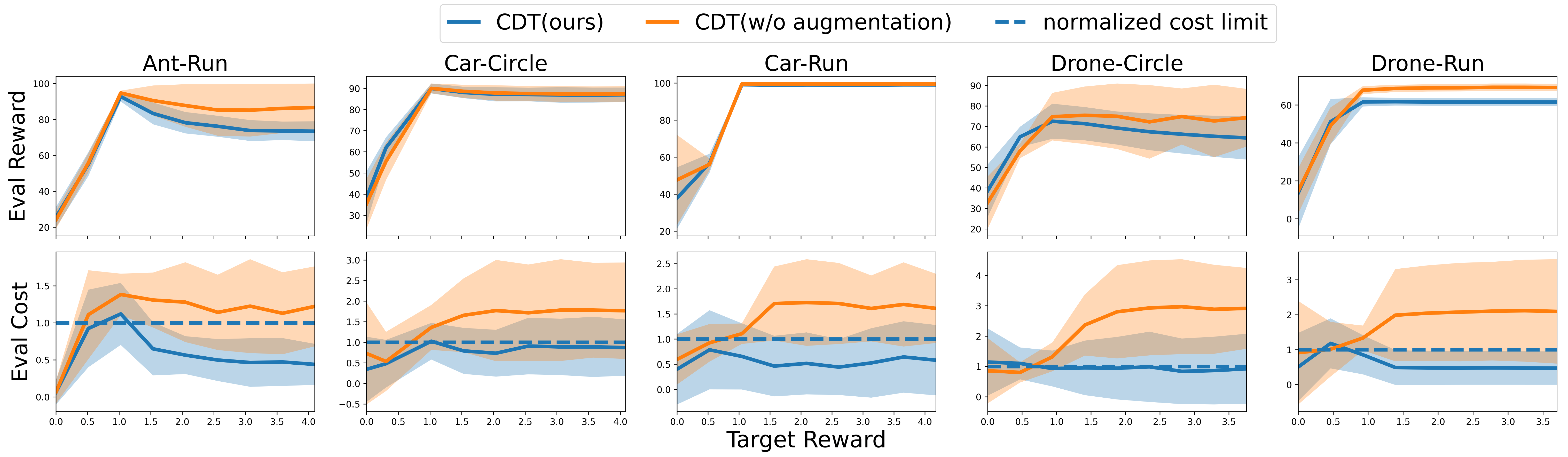}
\vspace{-4mm}
\caption{Ablation study of the data augmentation technique.
The x-axis is the target reward return.
The first row shows the evaluated reward, and the second row shows the evaluated cost under different target costs. 
The dashed line is the target cost threshold.}
\label{fig:exp-augment}
\vspace{-2mm}
\end{figure*}

\subsection{Can CDT achieve zero-shot adaption to different constraint thresholds?}
\label{sec:exp2}
As introduced in Sec.~\ref{sec:osrl-beyond-one-thres}, one significant advantage of CDT over baselines is its capability of zero-shot adaptation to different cost thresholds.
It is obvious that the baselines introduced previously lack this capability because they need a fixed pre-defined threshold to solve a constrained optimization problem. Adapting them to new constraint conditions requires re-training.
To this end, we add another baseline Multi-task Behavior Cloning (\textbf{MT-BC}) to compare the zero-shot adaptation performance with CDT~\cite{xu2022prompting}. 
We view each cost return threshold as a task and concatenate the task information (episodic cost return) to the corresponding task's states and train the agent via BC. Namely, the BC policy predicts an action that is conditioned on both state and cost threshold: $a_t = \pi_{\text{MT-BC}}(s_t,\kappa)$.

We fix the target reward and vary the target cost for evaluation rollouts to obtain the results in Fig. \ref{fig:zero-shot}. 
We can see that MT-BC has certain adaptation capabilities for the in-distribution target costs. 
However, when the cost limit exceeds the maximum cost in the datasets, the actual cost increases greatly (\texttt{Car-Run}), or the reward decreases significantly in other tasks.
On the contrary, the actual cost of CDT is strongly correlated with the target cost return and under the dashed threshold line, which shows great \textbf{interpolation} capability. 
The curves saturate at certain target costs that are beyond the maximum one in the training data, which shows that the agent can maintain safety even when performing \textbf{extrapolation} over unseen target cost returns. 
Furthermore, the actual reward return of CDT does not drop compared to MT-BC.

\subsection{Is CDT robust to conflict reward returns?}
% Compare with no augmentation.

As mentioned in section \ref{sec:cdt}, there exists infeasible target cost and reward pairs that could influence safety performance, which motivates us to propose the data augmentation technique.
To test its effectiveness, we fix the target cost return and vary the target reward to evaluate CDT and CDT without data augmentation. The results are shown in Fig. \ref{fig:exp-augment}.
We can observe that the actual reward increases and then saturates as the target reward increases.
However, the actual cost keeps increasing and finally exceeds the cost limit if removing data augmentation, while CDT can maintain safety even if the target return is large.
The results show that data augmentation is a necessary component in CDT to handle conflicting target returns.

\subsection{Can $\epsilon$-reducible characterize the task difficulty?}

To corroborate our hypothesis on the reducibility attribute of offline datasets in Sec. \ref{sec:offline-safe-rl}, we conduct experiments with both full (small $\epsilon$) and reduced (large $\epsilon$) datasets. The reduced dataset was constructed by removing trajectories whose costs exceeded the threshold and with high rewards. This process ensures that the most rewarding trajectories are safe, i.e., $\text{PF}(\kappa, \Tcal)>\text{IPF}(\kappa, \Tcal)$. 
Then we train standard offline RL algorithms, such as DT \cite{chen2021decision}, BCQ \cite{fujimoto2019off}, and BEAR \cite{kumar2019stabilizing} on these datasets. Due to the page limit, we present the results in Appendix \ref{app:data-reduction}.

The results show that these algorithms perform poorly in safety performance on the full dataset, which is as expected since maximizing reward is the sole objective. However, when training them on reduced datasets with large $\epsilon$ values, we can observe a significant improvement in terms of safety performance. 
This observation aligns with our conjecture: larger $\epsilon$-reducible problems are relatively easier to solve for the same task, as using standard offline RL algorithms can achieve good performance. 
% The results of these evaluations are detailed in Appendix \ref{app:data-reduction}.

\vspace{-1mm}
\section{Conclusion}
\vspace{-1mm}
We study the offline safe RL problem from the multi-objective optimization perspective and propose an empirically verified $\epsilon$-reducible concept to characterize the task difficulty. 
We further propose the CDT method that is capable of learning a safe and high-reward policy in challenging offline safe RL tasks.
More importantly, CDT can achieve zero-shot adaptation to different constraint thresholds without re-training and is robust to conflicting target returns, while prior works fail.
These advantages make CDT preferable for real-world applications with safety constraints.
% and we hope our findings can inspire more interdisciplinary research in this direction.
% We believe our work serves as useful guidance for future work.
% The potential negative societal impact and limitation of this work is that we can hardly guarantee safety of a trained policy without testing them in the environment, so how to provide rigorous theoretical analysis for offline safe RL is a promising and challenging future direction. 
% We hope our findings can inspire more interdisciplinary research in this direction, because safety is a nonnegligible factor in practical applications.
% The potential negative social impact is that the misuse of this work in safety-critical scenarios can cause unexpected damage.

There are also several limitations of CDT: 1) it more computing resources due to the Transformer architecture; 2) it lacks rigorous theoretical guarantees for safety; 3) it requires instant reward and cost feedback during the policy deployment and rollout; 4) improper target reward return and cost return can still deteriorate the performance, and 5) achieving zero-constraint violations is still challenging.
Therefore, studying a more lightweight method to address the above issues could be promising for future work. 
Nevertheless, we hope our findings can inspire more research in this direction to study the safety and generalization capability in offline learning.

\subsubsection*{Acknowledgements}
We express our profound gratitude to the National Science Foundation for their generous support under the CAREER CNS-2047454 grant. Funding to attend this conference was also provided by the CMU GSA/Provost Conference Funding. Equally, we extend our sincerest appreciation to the reviewers for their invaluable insights and constructive suggestions, which significantly contributed to the enhancement of our manuscript.

\clearpage

\bibliography{main}
\bibliographystyle{icml2023}

%%%%%%%%%%%%%%%%%%%%%%%%%%%%%%%%%%%%%%%%%%%%%%%%%%%%%%%%%%%%%%%%%%%%%%%%%%%%%%%
%%%%%%%%%%%%%%%%%%%%%%%%%%%%%%%%%%%%%%%%%%%%%%%%%%%%%%%%%%%%%%%%%%%%%%%%%%%%%%%
% APPENDIX
%%%%%%%%%%%%%%%%%%%%%%%%%%%%%%%%%%%%%%%%%%%%%%%%%%%%%%%%%%%%%%%%%%%%%%%%%%%%%%%
%%%%%%%%%%%%%%%%%%%%%%%%%%%%%%%%%%%%%%%%%%%%%%%%%%%%%%%%%%%%%%%%%%%%%%%%%%%%%%%
\newpage
\appendix
\onecolumn

\addcontentsline{toc}{section}{Appendix} % Add the appendix text to the document TOC

\part{} % Start the appendix part
\vspace{-1mm}
\parttoc % Insert the appendix TOC

\section{Results and Discussions on the $\epsilon$-reducible Datasets}
\begin{figure}[h]
    \centering
    \subfigure[Ant-Run task datasets, $\hat{\epsilon}|_{\kappa=10}=-0.040$ within full dataset, $\hat{\epsilon}|_{\kappa=10}=0.189$ within reduced dataset.]{
    \begin{minipage}{0.44\linewidth}
    \centering
    \includegraphics[width=1.0\linewidth]{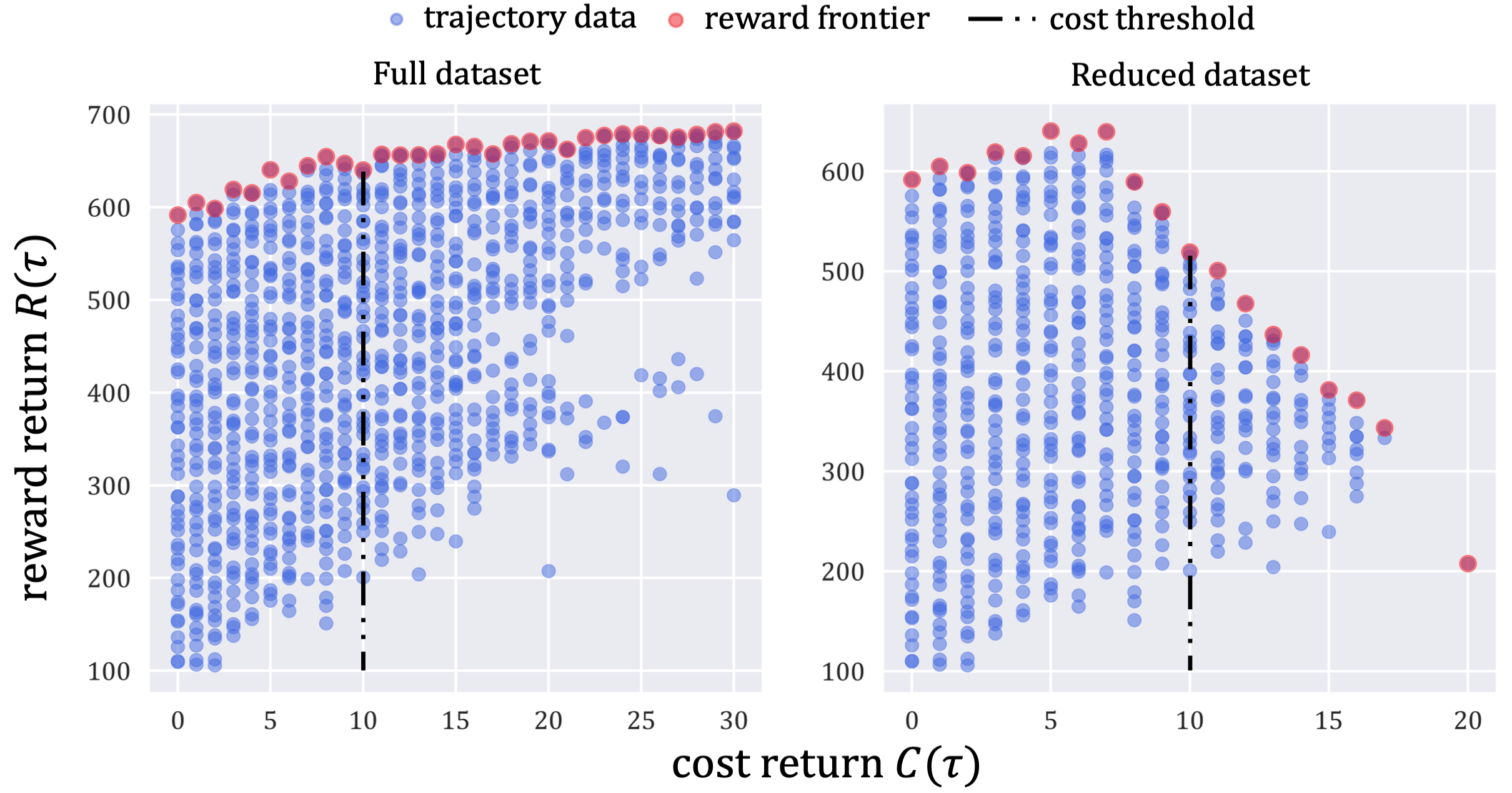}
    \end{minipage}%
    }%
    \ 
    \subfigure[Drone-Run task datasets, $\hat{\epsilon}|_{\kappa=10}=-0.281$ within full dataset, $\hat{\epsilon}|_{\kappa=10}=0.102$ within reduced dataset.]{
    \begin{minipage}{0.44\linewidth}
    \centering
    \includegraphics[width=1.0\linewidth]{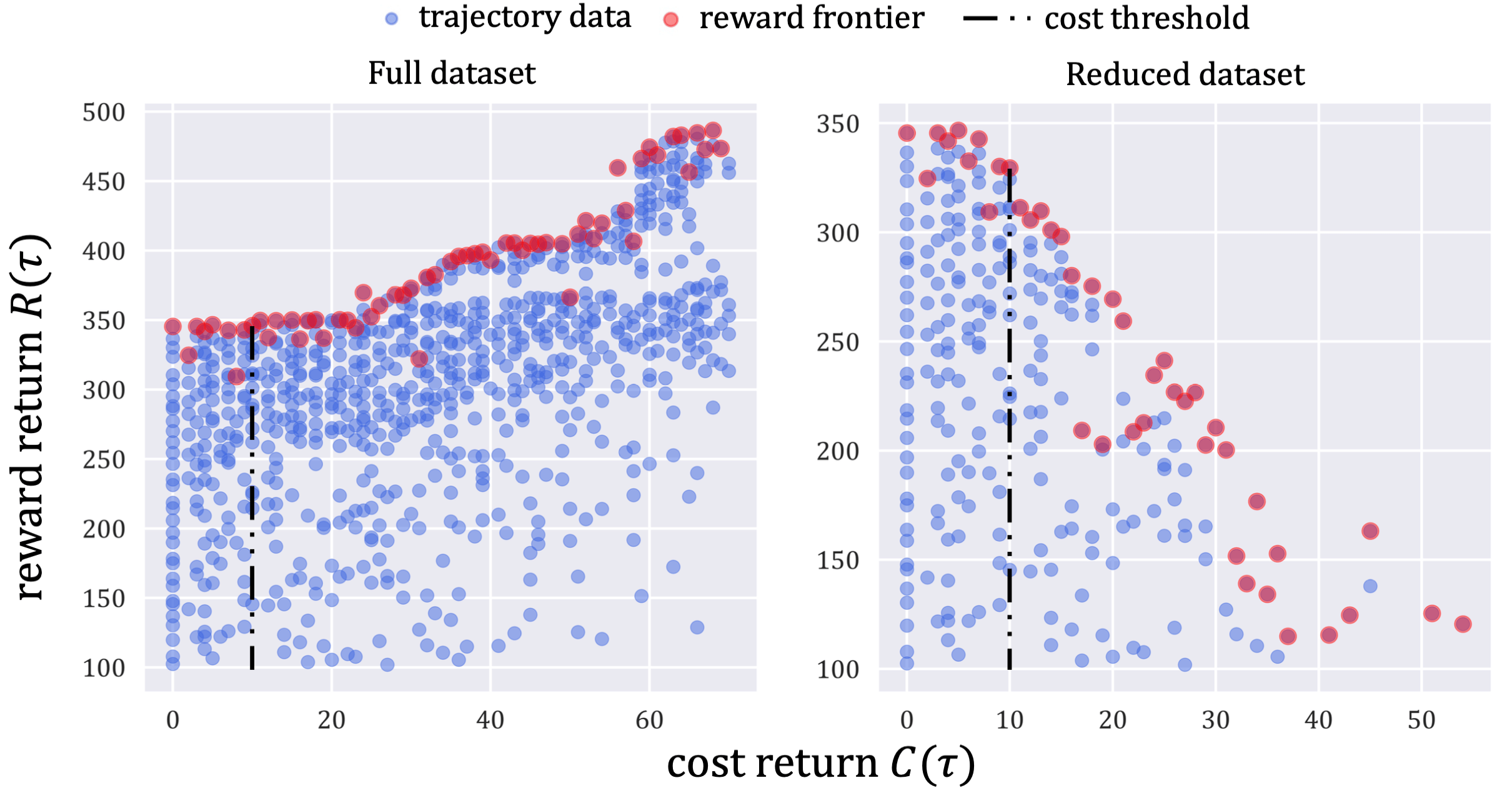}
    \end{minipage}%
    }%
    \centering
    % \vspace{-5mm}
    \caption{The cost-reward return plot of reduced datasets. The normalized $\epsilon$-reducible value for each dataset is normalized by the maximum return value in the dataset.}
    \label{fig:reduced-data}
\end{figure}

% \subsection{Experiment on $\epsilon$-reducible dataset}
\label{app:more-results}
To validate our hypothesis of the reducible property of the offline safe RL problem, we first construct reduced dataset (large $\epsilon$) from full dataset (small $\epsilon$) as shown in Figure \ref{fig:reduced-data}.
We remove the trajectories if their costs exceed the cost threshold and their rewards are higher than the trajectories whose costs are equal to the cost threshold to ensure that the trajectories with the highest reward satisfy the safety constraints, namely, $\text{PF}(\kappa, \Tcal)>\text{IPF}(\kappa, \Tcal).$
Then we train the standard offline RL methods such as DT \cite{chen2021decision}, BCQ \cite{fujimoto2019off} and BEAR \cite{kumar2019stabilizing} on both the full and reduced dataset. 
The evaluation results are shown in Figure \ref{tab:exp-reduced}.

As expected, on the small-$\epsilon$-reducible full data, these standard offline RL methods have poor safety performance since maximizing reward is the only optimization objective.
They learn to mimic the trajectories with higher rewards but violate the safety constraint in the dataset.
However, when they are trained on the large-$\epsilon$-reducible dataset, we can observe a significant improvement in terms of safety performance. 
The safety constraint will be naturally satisfied because the high-reward trajectories in the reduced dataset have smaller cost values than the threshold. 
In summary, we found that standard offline RL algorithms can solve the large-$\epsilon$-reducible problems well in most cases, which serves as strong evidence for our conjecture: the larger $\epsilon$, the easier to solve the problem.

\label{app:data-reduction}
% \subsection{Experiment results:}

\begin{table}[ht]
\centering
\renewcommand{\arraystretch}{1.1}
\resizebox{1.\linewidth}{!}{
\begin{tabular}{c|cccccc|cccccc}
% \hline
\toprule
                          & \multicolumn{6}{c|}{Full Dataset (small $\epsilon$)}                                                                                                                                                                                       & \multicolumn{6}{c}{Reduced Dataset (large $\epsilon$)}                                                                                                                                    \\ \cline{2-13} 
                          & \multicolumn{2}{c|}{Ant-Run ($\hat{\epsilon}=-0.040$) }                                                    & \multicolumn{2}{c|}{Drone-Run ($\hat{\epsilon}=-0.281$)}                                                  & \multicolumn{2}{c|}{Average}                               & \multicolumn{2}{c|}{Ant-Run ($\hat{\epsilon}=0.189$)}                        & \multicolumn{2}{c|}{Drone-Run ($\hat{\epsilon}=0.102$)}                                                  & \multicolumn{2}{c}{Average}    \\
\multirow{-3}{*}{Methods} & Reward $\uparrow$                      & \multicolumn{1}{c|}{Cost $\downarrow$}                        & Reward $\uparrow$                      & \multicolumn{1}{c|}{Cost $\downarrow$}                        & Reward $\uparrow$                      & Cost $\downarrow$                       & Reward $\uparrow$        & \multicolumn{1}{c|}{Cost $\downarrow$}          & Reward $\uparrow$                      & \multicolumn{1}{c|}{Cost $\downarrow$}                        & Reward $\uparrow$        & Cost $\downarrow$         \\ \hline
DT                        & {\color[HTML]{656565} 96.13} & \multicolumn{1}{c|}{{\color[HTML]{656565} 2.47}} & {\color[HTML]{656565} 49.09} & \multicolumn{1}{c|}{{\color[HTML]{656565} 4.69}} & {\color[HTML]{656565} 72.61} & {\color[HTML]{656565} 3.58} & \textbf{81.26} & \multicolumn{1}{c|}{\textbf{0.91}} & {\color[HTML]{656565} 63.88} & \multicolumn{1}{c|}{{\color[HTML]{656565} 1.03}} & \textbf{72.57} & \textbf{0.97} \\
BCQ                       & {\color[HTML]{656565} 97.99} & \multicolumn{1}{c|}{{\color[HTML]{656565} 5.39}} & {\color[HTML]{656565} 60.1}  & \multicolumn{1}{c|}{{\color[HTML]{656565} 3.01}} & {\color[HTML]{656565} 79.04} & {\color[HTML]{656565} 4.2}  & 82.59          & \multicolumn{1}{c|}{1.19}          & \textbf{42.34}               & \multicolumn{1}{c|}{\textbf{0.51}}               & \textbf{62.46} & \textbf{0.85} \\
BEAR                      & {\color[HTML]{656565} 91.05} & \multicolumn{1}{c|}{{\color[HTML]{656565} 1.69}} & \textbf{42.13}               & \multicolumn{1}{c|}{\textbf{0.48}}               & {\color[HTML]{656565} 66.59} & {\color[HTML]{656565} 1.08} & \textbf{84.2}  & \multicolumn{1}{c|}{\textbf{0.55}} & \textbf{33.29}               & \multicolumn{1}{c|}{\textbf{0.0}}                & \textbf{58.74} & \textbf{0.28} \\ 
\bottomrule
% \hline
\end{tabular}
}
\caption{Evaluation results of the normalized reward and cost. The cost threshold is 1.
$\uparrow$ / $\downarrow$: the higher/lower, the better.
Each value is averaged over 20 episodes and 3 seeds.
\textbf{Bold}: Safe agents whose normalized cost is smaller than 1. 
{\color[HTML]{656565} Gray}: Unsafe agents. The normalized $\epsilon$-reducible value for each dataset, which is normalized by the maximum return value in the dataset, is also labeled in the table.
}
\label{tab:exp-reduced}
\end{table}

% \subsection{Further discussion about $\epsilon$-reducible}
% \label{app:proof}
One implicit assumption for $\epsilon$-reducible property is that the learning capability of an offline RL learner is limited, and the datasets are of good quality that can cover high-reward spaces, i.e., the agent can hardly achieve any trajectory $\tau$ with a higher reward $r > \text{PF}(\Tcal, \kappa)$ under safety constraint $\kappa$ given the collected dataset $\Tcal$. 
With this limited learning ability assumption, given a positive-reducible dataset, the agent will not achieve a reward that is higher than $\text{PF}(\Tcal, \kappa)$ even if we remove safety constraints during offline training.

\clearpage
\section{Implementation Details}
The implementation follows the CORL \cite{tarasov2022corl} package and is available at \url{https://github.com/liuzuxin/OSRL}.
\label{app:implement}
% The CDT training process is shown in Alg.~\ref{algo:training}.
\subsection{CDT Training Procedure Pseudo Code}
\begin{algorithm}[h]
\caption{CDT Training Procedure}
{\bfseries Input:} \raggedright Transformer model $\pi_\theta$, dataset $\Tcal$, learning rate $\alpha$, context length $K$, batch size $B$, entropy weight $\lambda$, gradient steps $M$, maximum episode length $T$, augment samples $N$, reward sample max $r_{max}$ \par
{\bfseries Output:} \raggedright trained Transformer model $\pi_\theta$ \par
\begin{algorithmic}[1] 
\STATE Augment dataset: $\Tcal \leftarrow Augment(\Tcal, N, r_{max})$
\FOR{$\text{update step}=1,..., M$}
\STATE $\triangleright$ \textit{sample a batch of sequences of length $K$}
\STATE $ \mathcal{B} = \{\mathbf{a}_{i, t}, \mathbf{o}_{i, t} \}_{i=1}^B \sim \Tcal, t \sim \text{SampleInt}(1, T)$
\STATE $\triangleright$ \textit{compute the NLL loss and entropy loss}
\STATE $\ell_{\text{nll}} = - \frac{1}{|\mathcal{B}|} \sum_{\mathbf{a}, \mathbf{o} \in \mathcal{B}}\log \pi_\theta (\mathbf{a} | \mathbf{o}) $
\STATE $\ell_{\text{ent}} = - \frac{1}{|\mathcal{B}|} \sum_{\mathbf{o} \in \mathcal{B}} H[\pi_\theta(\cdot | \mathbf{o})]  $
\STATE $\triangleright$ \textit{update the policy parameter}
\STATE $\ell_{\text{cdt}} =  \ell_{\text{nll}} + \lambda \ell_{\text{ent}}$
\STATE $\theta \leftarrow  \theta - \alpha \nabla_\theta\ell_{\text{cdt}}$
\ENDFOR
\end{algorithmic} \label{algo:training}
\end{algorithm}

\subsection{Dataset Collection}
\label{app:data-collection}
\subsubsection{Algorithm}
We collect offline datasets using the CPPO safe RL approach~\cite{stooke2020responsive}, which is an improved version of the PPO-Lagrangian method by using a PID controller to update the dual variable~\cite{ray2019benchmarking}. Suppose the reward and cost value functions $V_{r}, V_{c}$ are parameterized by $\theta_r$ and $\theta_c$ networks respectively. We use GAE~\cite{schulman2015high} to update the value functions:
\vspace*{-1mm}
\begin{equation}
\begin{aligned}
    \theta_r \leftarrow \arg\min_{\theta_r} \mathbb{E}_{s_{\tau}\sim \mathcal{D}} \left[\left(V_r(s_{\tau}) - \sum_{t=0}^{T-\tau} (\lambda^{gae}\gamma)^t r(s_t,a_t) \right)^2\right] \\
    \theta_c \leftarrow \arg\min_{\theta_c} \mathbb{E}_{s_{\tau}\sim \mathcal{D}} \left[\left(V_c(s_{\tau}) - \sum_{t=0}^{T-\tau} (\lambda^{gae}\gamma)^t c(s_t,a_t) \right)^2\right]
\end{aligned}
\label{eq:ppo-value}
\end{equation}
where $\gamma$ is the discounting factor, and $\lambda^{gae}$ is the GAE constant.
% To compute the policy loss, we first calculate the importance sampling ratio for a given state-action pair $r_{IS}(s,a)=\frac{\pi_{\phi}(a|s)}{\pi_i(a|s)}$ and then the surrogate loss is 
% \begin{equation}
% \begin{aligned}
%     \mathcal{L}_r= {\mathbb{E}}_{\mathcal{D}}\left[\min \left(r_{IS} A_r, \operatorname{clip}\left(r_{IS}, 1-\epsilon, 1+\epsilon\right) A_r\right)\right],\\
%     \mathcal{L}_c= {\mathbb{E}}_{\mathcal{D}}\left[\max \left(r_{IS} A_c, \operatorname{clip}\left(r_{IS}, 1-\epsilon, 1+\epsilon\right) A_c\right)\right]
% \end{aligned}
% \label{eq:ppo-surr}
% \end{equation}
% where $\epsilon$ is the clip ratio and $A$ is advantage function: $A_r(s,a) = r(s,a) + \gamma V(s') - V(s)$. 
The objective of clipped PPO has the form \citep{schulman2017proximal}:
\vspace*{-1mm}
\begin{equation}
    \ell_{ppo} = \min (\frac{\pi_{\theta}(a|s)}{\pi_{\theta_k}(a|s)} A^{\pi_{\theta_k}}_r(s, a), \text{clip}(\frac{\pi_{\theta}(a|s)}{\pi_{\theta_k}(a|s)}, 1-\epsilon_{\text{clip}}, 1+\epsilon_{\text{clip}})A^{\pi_{\theta_k}}_r(s, a) )
\end{equation}
% \vspace*{-3mm}
We use PID Lagrangian \cite{stooke2020responsive} that addresses the oscillation and overshoot problem in Lagrangian methods. The loss of the PPO-Lagrangian has the form:
\vspace*{-3mm}
\begin{equation}
    \ell_{ppol} = \frac{1}{1+\lambda} ( \ell_{ppo} - \lambda A^{\pi_{\theta_k}}_c(s, a))
    \label{eq:lossppol}
\end{equation}
The Lagrangian multiplier $\lambda$ is computed by applying feedback control to $V_c^{\pi}$ and is determined by positive constants $k_P$, $k_I$, and $k_D$. Instead of using a fixed cost threshold $\epsilon$, we apply a time-varying cost threshold so that we can collect data within a wide range of reward and cost values. The procedure of 
 \texttt{CPPO} is summarized in Alg.~\ref{algo: threshold-varying ppo-lag}. 

\begin{algorithm}[htb]
\caption{CPPO}
{\bfseries Input:} \raggedright Cost interval $[C_{\text{min}}, C_{\text{max}}]$, starting epoch $n_1$, ending epoch $n_2$, epoch $n$, PID parameter $\{k_P, k_I, k_D\}$, learning rate $\eta_\phi$. \par
{\bfseries Output:} \raggedright Policy parameter $\phi$, dataset $\Gamma$. \par
\begin{algorithmic}[1] % The number tells where the line numbering should start
\STATE Initialization: target cost error $e_0 \leftarrow 0$, replay buffer $\mathcal{D} \leftarrow \{\}$, dataset $\Gamma \leftarrow \{\}$.
\FOR{$i=1,..., n$}
\STATE Compute the threshold $\epsilon\leftarrow\min \{ C_{\text{max}}, \max \{C_{\text{min}}, \frac{C_{\text{max}} - C_{\text{min}}}{n_2 - n_1} (i-n_1) + \epsilon_1\} \}$
\STATE Sample $N$ trajectories $\{s_0, a_0, \dots, s_T\}_{n=1,...,N}$ with policy $\pi_i$ and store transition data to replay buffer $\mathcal{D}$.
\STATE Compute the expectation of cost return $J_c$.
\STATE Calculate the error between the real cost and the cost threshold: $e_i \leftarrow J_c - \epsilon$.
\STATE Update dual variable $\lambda \leftarrow k_P \ e_i + k_I \max\{0,\sum_{j=1}^{i} e_j\} + k_D \max\{0, e_i - e_{i-1}\} $.
\STATE Update value functions based on Eq. (\ref{eq:ppo-value}).\\
\STATE Update policy: $\phi\leftarrow \phi + \eta_{\phi} \nabla_\phi (\mathcal{L}_r - \lambda \mathcal{L}_c)$.
\STATE Save the dataset $\Gamma \leftarrow \Gamma \cup \mathcal{D}$ and empty the buffer $\mathcal{D}\leftarrow \{\}$.
\ENDFOR
\end{algorithmic} \label{algo: threshold-varying ppo-lag}
\end{algorithm}

\subsubsection{Hyperparameters of the expert CPPO policy}
To collect datasets with a large range of cost and reward return values, we fine-tune the hyperparameters in  \texttt{CPPO}. The key hyperparameters for our dataset collection are listed in Tab.~\ref{tab:dataset-hyperparameters}.
\begin{table}[htb]
    \centering

    \begin{tabular}{cccccc}
       \toprule
       parameters & Car-Run & Ant-Run & Car-Circle & Drone-Circle & Drone-Run \\
       \midrule
       % $\eta_\phi$ & $3\times10^{-4}$ & $3\times10^{-4}$ & $3\times10^{-4}$ & $3\times10^{-4}$ & $3\times10^{-4}$ \\
       $\epsilon_{\text{clip}}$ & $0.2$ & $0.2$ & $0.2$ & $0.15$ & $0.15$ \\
       $\lambda^{\text{gae}}$ & $0.97$ & $0.97$ & $0.97$ & $0.95$ & $0.95$ \\
       $\gamma$ & $0.99$ & $0.99$ & $0.99$ & $0.98$ & $0.98$ \\
       % $[k_P, k_I, k_D]$ & $[0.1, 0.003, 0.001]$ & $[0.1, 0.003, 0.001]$ & $[0.1, 0.003, 0.001]$ & $[0.1, 0.003, 0.001]$ & $[0.1, 0.003, 0.001]$ \\
       $[\epsilon_1, \epsilon_2]$ & $[5, 80]$ & $[5, 80]$ & $[5, 80]$ & $[10, 80]$ & $[5, 80]$ \\
       $[n_1, n_2]$ & $[50, 400]$ & $[45, 200]$ & $[50, 200]$ & $[20, 550]$ & $[10, 150]$ \\
       $n$ & $400$ & $210$ & $210$ & $570$ & $160$ \\
       \bottomrule
    \end{tabular}
    \caption{The hyperparameters for \texttt{CPPO} algorithm for data collection}

    \label{tab:dataset-hyperparameters}
\end{table}

\subsubsection{Dataset visualization}
\label{section: RF curves}
\begin{figure}[ht]
% \vspace*{-2mm}
\centering
\includegraphics[width=0.95\linewidth]{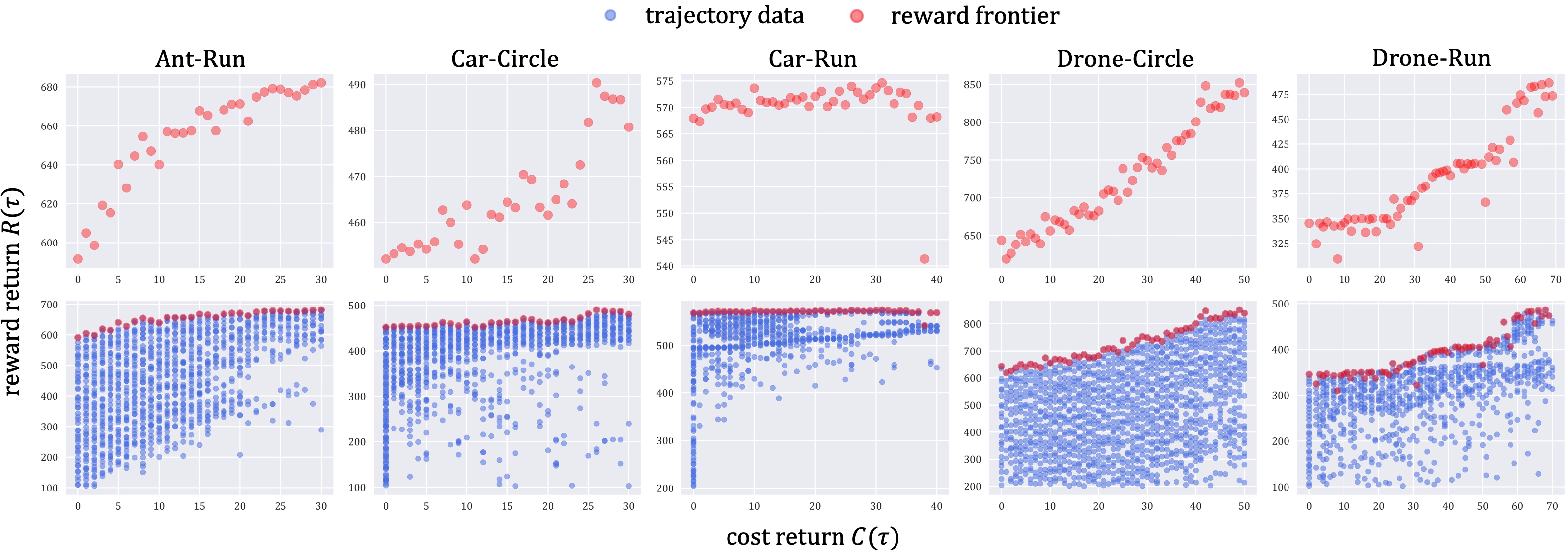}
\vspace*{-4mm}
\caption{Cost-reward return plot. The reward frontiers of each sampling cost are marked in red. Each column represents a task. The first row shows the reward frontier points in the dataset, and the second row shows the whole dataset. Each point represents collected trajectories (not necessarily to be unique) with corresponding episodic cost and reward value. The cost values are discrete because all these tasks adopt the $0$-$1$ cost. The normalized $\epsilon$-reducible values at threshold $\kappa=10$ for these datasets are listed as Ant-Run: -0.040, Car-Circle: -0.056, Car-Run: -0.005, Drone-Circle: -0.208, Drone-Run: -0.281. The normalized $\epsilon$-reducible value for each dataset is normalized by the maximum return value in the dataset.}
\label{fig:dataset return}
% \vspace*{-4mm}
\end{figure}

The dataset cost-reward return plots for the training tasks \texttt{Ant-Run}, \texttt{Car-Circle}, \texttt{Car-Run}, \texttt{Drone-Circle}, and \texttt{Drone-Run} are shown in Fig.~\ref{fig:dataset return}. Due to the limited sampling number, the reward frontier value is not monotonically increasing with respect to the cost. However, we can observe the trend that high-cost values are with high maximum reward values in most cases. This is consistent with our intuition and our motivation for loosening safety constraints: large reward values are traded off by the high risk of violating safety constraints. However, in some cases, such as the \texttt{Car-Run} task, this trend is not conspicuous. It is because \texttt{Car-Run} is an easy task -- with easy robot dynamics and a simple environment, that the safety constraint can hardly block the \texttt{CPPO} agent from reaching a higher reward.

\section{Experiment Setting and Hyperparameters}

\subsection{Experiment Description}
We use the Bullet safety gym~\citep{gronauer2022bullet} environments for this set of experiments. 
In the Run tasks, agents are rewarded for running fast between two safety boundaries and are given costs for violation constraints if they run across the boundaries or exceed an agent-specific velocity threshold.
The reward and cost functions are defined as:
\begin{align*}
    % r(s) & = \sqrt{(x_{t-1}-g_x)^2 + (y_{t-1}-g_y)^2} - \sqrt{(x_t-g_x)^2 + (y_t-g_y)^2} + r_{robot}(s) \\
    r(\bm{s_t}) & = ||\bm{x_{t-1}}-\bm{g}||_2 - ||\bm{x}_t-\bm{g}||_2 + r_{robot}(s_t) \\
    c(\bm{s_t}) & = \bm{1} (|y| > y_{lim}) + \bm{1}(||\bm{v_t}||_2 > v_{lim})
\end{align*}
where $v_{lim}$ is the speed limit, $y_{lim}$ specifies the safety region, $\bm{v_t} = [v_x, v_y]$ is the velocity of the agent at timestamp $t$, $\bm{g}= [g_x, g_y]$ is the position of a fictitious target, $\bm{x_t} = [x_t, y_t]$ is the position of the agent at timestamp $t$, and $r_{robot}(\bm{s_t})$ is the specific reward for different robot. 
For example, an ant robot will gain reward if its feet do not collide with each other. 
In the Circle tasks, the agents are rewarded for running in a circle in a clockwise direction but are constrained to stay within a safe region that is smaller than the radius of the target circle.
The reward and cost functions are defined as:
\begin{align*}
    r(\bm{s_t}) & = \frac{-y_t v_x + x_t v_y}{1 + | ||\bm{x_t}||_2-r|} + r_{robot}(\bm{s_t}) \\
    c(\bm{s_t}) & = \bm{1}(|x| > x_{lim})
\end{align*}
where $r$ is the radius of the circle, and $x_{lim}$ specifies the range of the safety region.

\subsection{Hyperparameters}
\label{app:exp-hyperparameters}
For baselines, we use Gaussian policies with mean vectors given as the outputs of neural networks, and with variances that are separate learnable parameters. 
The policy networks and Q networks for all experiments have two hidden layers with ReLU activation functions. 
The $K_P, K_I$ and $K_D$ are the PID parameters \cite{stooke2020responsive} that control the Lagrangian multiplier for the Lagrangian-based algorithms.
We use the same $10^5$ gradient steps and rollout length which is the maximum episode length for CDT and baselines for fair comparison.
The cost threshold for baselines is 10 across all the tasks.
The hyperparameters that are not mentioned are in their default value for baselines.
The complete hyperparameters used in the experiments are shown in Table \ref{tab:exp-parameters}. 

\begin{table}[h]
\centering
\renewcommand{\arraystretch}{1.1}
\resizebox{1.\linewidth}{!}{
\begin{tabular}{ccccccccc}
\cline{1-2} \cline{4-9}
Parameter                 & All tasks    &  & Parameter                          & Ant-Run & Car-Circle & Car-Run & Drone-Circle & Drone-Run \\ \cline{1-2} \cline{4-9} 
Number of layers          & 3            &  & \multirow{2}{*}{Actor hidden size} & \multicolumn{5}{c}{[256, 256] BCQ-Lag, BEAR-Lag}          \\
Number of attention heads & 8            &  &                                    & \multicolumn{5}{c}{[300, 300] CPQ}                        \\
Embedding dimension       & 128          &  & \multirow{2}{*}{VAE hidden size}   & \multicolumn{5}{c}{[750, 750] BEAR-Lag}                   \\
Batch size                & 2048         &  &                                    & \multicolumn{5}{c}{[400, 400] CPQ}                        \\
Context length $K$        & 10           &  & Rollout length                     & 200     & 300        & 200     & 300          & 100       \\
Learning rate             & 0.0001       &  & [$K_P, K_I, K_D$]                  & \multicolumn{5}{c}{[0.1, 0.003, 0.001] BCQ-Lag, BEAR-Lag} \\
Droupout                  & 0.1          &  & Batch size                         & \multicolumn{5}{c}{512}                                   \\
Adam betas                & (0.9, 0.999) &  & Actor learning rate                & 0.0001  & 0.001      & 0.0001  & 0.0001       & 0.001     \\
Grad norm clip            & 0.25         &  & Critic learning rate               & 0.001   & 0.001      & 0.001   & 0.001        & 0.001     \\ \cline{1-2} \cline{4-9}
\end{tabular}
}
\caption{Hyperparameters for CDT (left) and baselines (right).}
\label{tab:exp-parameters}
\end{table}

% \section{More Experiment Results}

% \subsection{Context length}
% To assess the influence of access to previous states, actions, and returns, we ablate on the context length $K$. 
% As shown in Figure \ref{fig:exp-seqlen}, CDT can learn safe policies when given a limited memory of the past information. 
% Besides, the task performance of CDT is less sensitive to the context length compared to its safety performance.
% Large context length should be avoided since it would cause the training unstable and over-fitting. 

% \begin{figure}[h]
%     \centering
%     \includegraphics[width=0.8\linewidth]{figures/exp-seqlen.png}
%     \vspace{-5mm}
%     \caption{Effect of the context length $K$.}
%     \label{fig:exp-seqlen}
% \end{figure}

% \subsection{}

\clearpage
\section{More Results and Discussions}

\subsection{Why do stochastic policies have better empirical safety performance?}
\label{app:stochastic-policy}

From the experiments, we can observe that using a stochastic policy head is much better than using a deterministic policy.
We conjecture that this is because stochastic policies can help alleviate potential constraint violations caused by out-of-distribution (OOD) actions and extrapolation errors in offline safe RL. In other words, if the policy learns to incorrectly estimate an unseen action as the most rewarding and safe, but it is actually unsafe, a deterministic policy will always take that action with probability 1. However, a stochastic policy can choose other safe actions by sampling, which improves safety.

We provide a numerical example as follows. Consider a problem with discrete action spaces: $\mathcal{A} = \{a_1, a_2, a_3\}$. We focus on the decision at state $s$.
The rewards and costs are 
$$r(s,a_1) = 1, r(s,a_2)=2, r(s, a_3) = 3,$$ 
$$c(s,a_1) = 0, c(s,a_2)=0, c(s, a_3) = 10,$$ 
and the state-action pairs $(s, a_1), (s, a_2)$ are in the dataset while  $(s, a_3)$ is not.
Due to the increasing trend of reward and the safe actions based on the observed data at state $s$, the policy may wrongly estimates that the OOD action $a_3$ is most rewarding and also safe. 

Therefore, a deterministic policy will execute $a_3$ with probability 1, and the expected cost is 10.

In contrast, a stochastic policy may output a policy distribution that is proportional to the reward: $\pi(a_1|s) = 1/6, \pi(a_2|s)=2/6, \pi(a_3|s)=3/6$. Thus, the expected cost should be $\pi(a_3|s) c(s, a_3) = 3/6 \times 10 = 5$, which is significantly smaller than the deterministic policy's cost.

\begin{figure}[h]
    \centering
    \subfigure[When offline data is limited, we may not estimate the cost function correctly for out-of-distribution regions.]{
    \begin{minipage}{0.44\linewidth}
    \centering
    \includegraphics[width=1.0\linewidth]{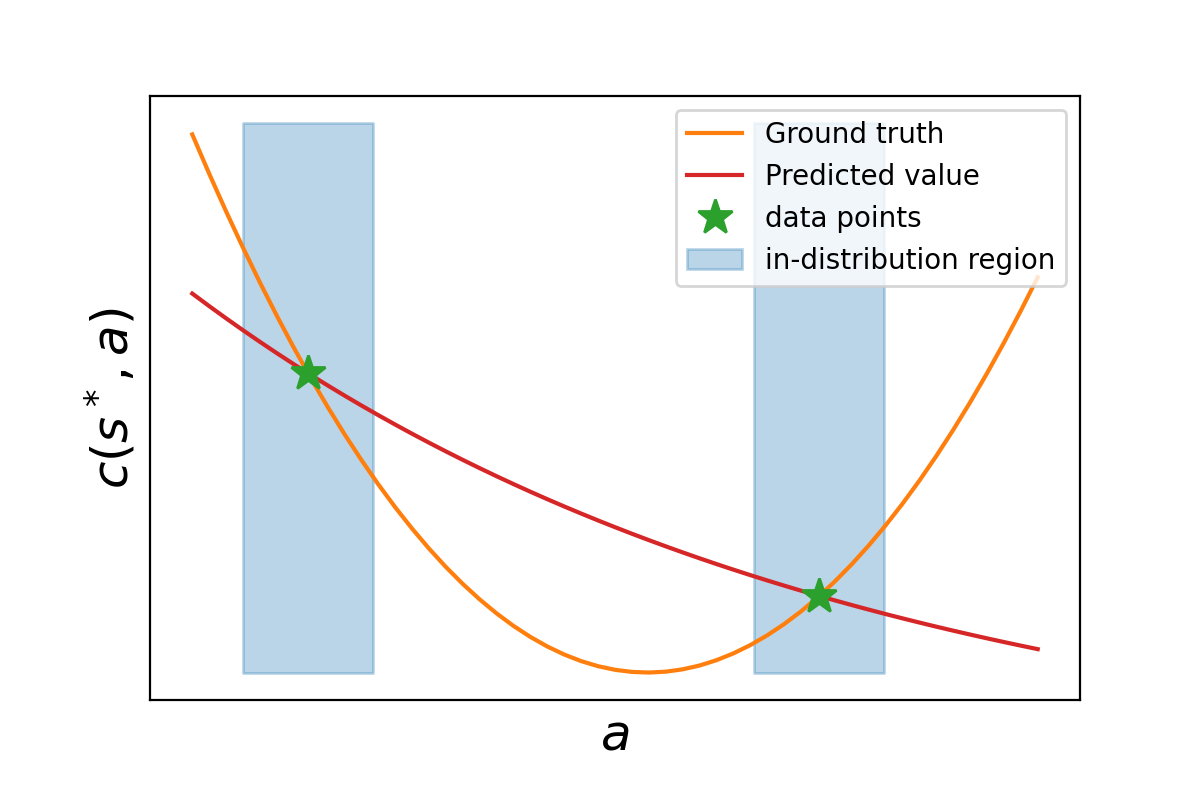}
    \end{minipage}%
    }%
    \ 
    \subfigure[The action from corresponding deterministic policy is always out-of-distribution while the stochastic policy can still sample in-distribution actions.]{
    \begin{minipage}{0.44\linewidth}
    \centering
    \includegraphics[width=1.0\linewidth]{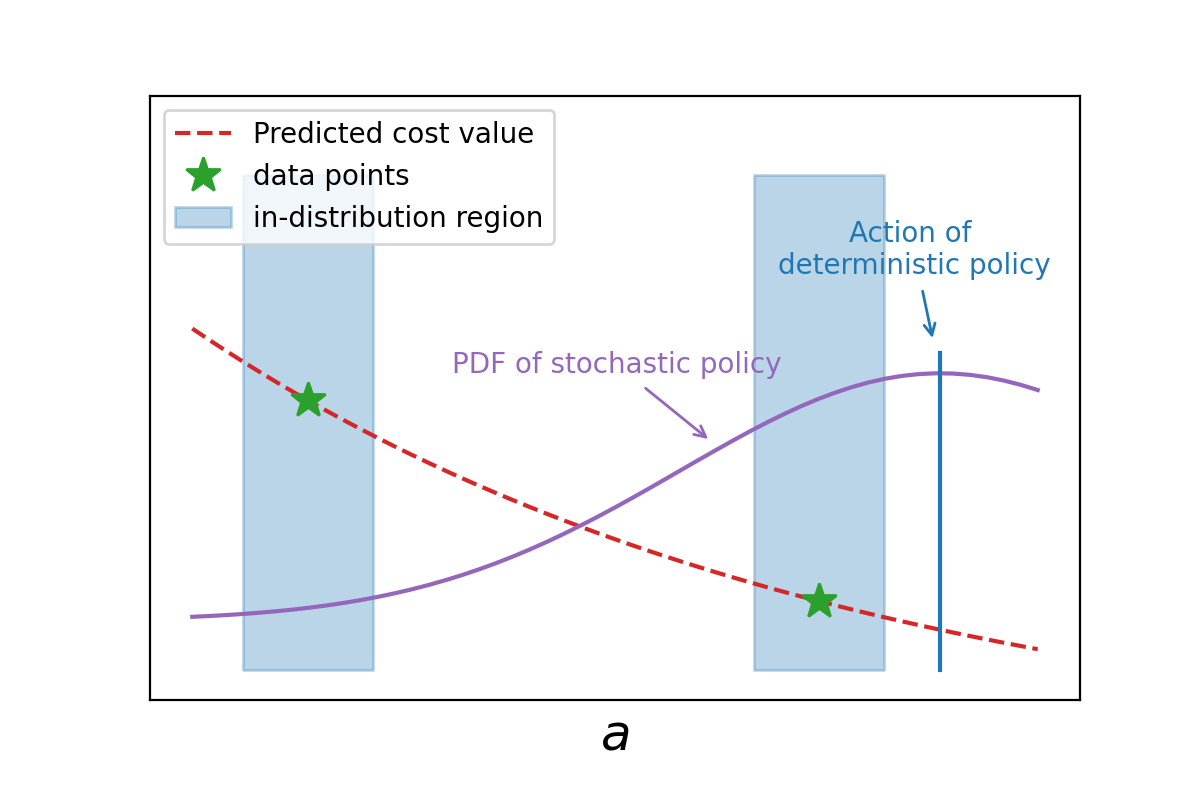}
    \end{minipage}%
    }%
    \centering
    % \vspace{-5mm}
    \caption{The predicted cost function and corresponding stochastic and deterministic policies. Deterministic policy usually suffers from out-of-distribution issue more severely, which leads to worse safety performance especially when the extrapolation error of cost estimation is large.}
    \label{fig:rebuttal_stochastic}
\end{figure}

We also provide a figure illustration for continuous action space.
As shown in fig.~\ref{fig:rebuttal_stochastic}, if we consider the cost function (which is also applicable to other safety-related functions depending on different methods) at a fixed state $s=s^*$, it may not be well-estimated with limited offline data. In this case, to reduce the cost to keep safety, the deterministic policy will output an out-of-distribution action (blue line), while a stochastic policy can still maintain a part of probability weight on in-distribution and safe regions (orange line).

Therefore, employing a stochastic policy can improve safety performance when facing out-of-distribution data and extrapolation errors, making it a crucial technique for CDT in offline safe RL. More in-depth theoretical analysis will be a promising direction for future research.
%We hope that the illustrative examples provided above and our explanations are helpful to the readers.

\clearpage
\subsection{Ablation studies for noisy datasets}
\label{app:noisy-dataset}

The real-world dataset could be noisy and contains trajectories that accidentally record high reward returns and small cost returns despite following a poor behavioral policy. 
Such lucky trajectories would be rare (outliers) and should be omitted in learning. However, when such a lucky trajectory exists in the dataset, it may affect the data augmentation procedure in CDT and thus negatively affect its performance. 
To investigate this issue, our implementation adopts two techniques. They are summarized as follows.

\begin{enumerate}
    \item The first technique is to associate each augmented return pair $(r, c)$ with a trajectory sampled in the neighborhood of the nearest and safe Pareto frontier data point, and the sampling distribution is based on a distance metric $W((r_p, c_p), (r', c')):=||(r_p, c_p)-(r', c')||_2+\beta$, where $\beta \in R_+$ is a constant, $(r_p, c_p)$ is the return of the Pareto frontier, and $(r', c') \in U((r_p, c_p))$ is the return of the data point in the neighbor of $(r_p, c_p)$. That is to say, we may associate the return pair $(r, c)$ to a range of data points around the Pareto frontier $(r_p, c_p)$. The probability of data selection is inversely proportional to the distance, i.e., $p((r, c) \leftarrow (r', c')) \propto 1 / {W((r_p, c_p), (r', c'))}$. This sampling-based association can not only increase the diversity of augmented trajectories, but also mitigate the outlier issue. 
    \item The second technique uses a density filter to remove such outliers with abnormal reward and cost returns when creating the training dataset. In particular, we implemented a grid filter that first segments the reward and cost return space into evenly distributed grids and then counts the number of trajectories within each grid. As such, we can easily filter the outlier trajectories of small densities.
\end{enumerate}

\begin{figure}[ht]
% \vspace*{-2mm}
\centering
\includegraphics[width=0.95\linewidth]{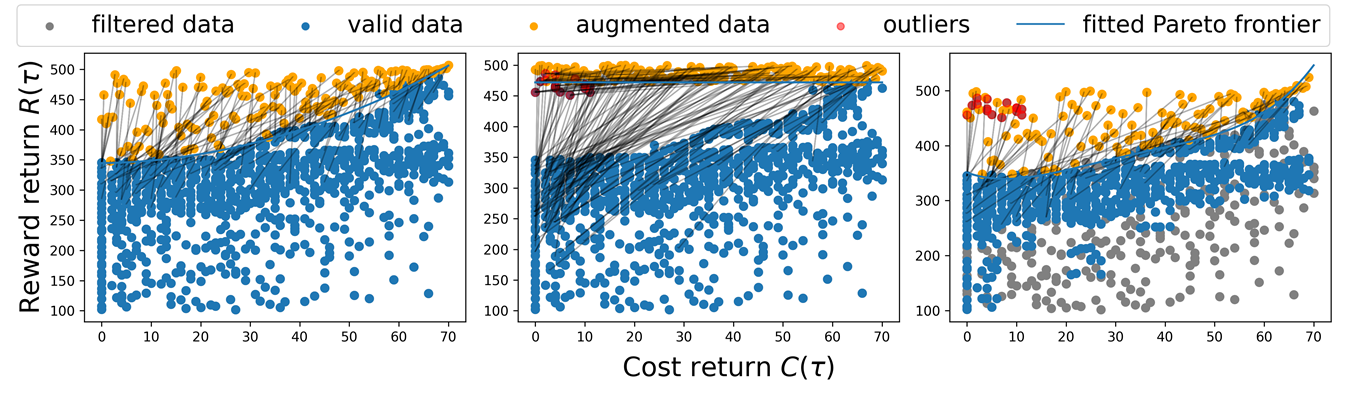}
\vspace*{-4mm}
\caption{Cost-reward return illustrations of the Drone-Run dataset. Each point denotes the trajectories with corresponding cost-return values. Left: the original dataset w/o outlier trajectories. Middle: sampling-based association with 2\% outlier trajectories. Right: density filter with 2\% outlier trajectories. The black lines show the association.}
\label{fig:outliers}
% \vspace*{-4mm}
\end{figure}

To show the effectiveness of the above techniques, we create such a dataset that contains different portions ($\alpha \% = 0.1\%, 0.4\%,1\%,1.5\%,2\%$) of outlier trajectories based on the Drone-Run dataset. 
We consider the task with stochastic reward and cost function, i.e., high-cost trajectories have the probability of $\alpha\%$ to be labeled as a ``lucky" trajectory with high reward and low cost. 
Specifically, we select $\alpha \%$ high-cost trajectories and modify their cost return to be less than the cost threshold, as shown in the red dots in Fig.~\ref{fig:outliers}. 

The middle figure visualizes the sampling-based association technique. Note that the outliers do not only affect the association, but also influence the sampled return and cost pairs. We can see that some of the augmented returns are wrongly associated with the outliers, but the remaining ones are paired correctly.

The right figure demonstrates the result of applying a density filter. We can see that the outliers are removed since their densities on the reward and cost return space are small. Therefore, the augmented returns are associated correctly. However, we can observe that some normal trajectories on the lower right regions are also filtered. 

We perform CDT to train the agents based on the above two techniques. The evaluation results are listed in Table \ref{tab:rebuttal-outliers}. Note that the natural performance of CDT on the clean dataset is $r=63.64, c=0.58$.
We can observe that as the outlier percentage increases, CDT's safety performance does be affected: the cost may also increase. However, in most cases, CDT can still learn a safe policy, and the reward doesn't drop too much. 
The results indicate that the proposed two techniques can mitigate the negative effect induced by outlier trajectories.

\begin{table*}[ht]
\centering
\renewcommand{\arraystretch}{1.1}
\resizebox{1.\linewidth}{!}{
% \begin{tabular}{c|cc|cc|cc|cc|cc}
\begin{tabular}{c|cc|cc|cc|cc|cc|cc}
% \hline
\toprule
                          & \multicolumn{2}{c|}{$\alpha = 0.1$}                                                  & \multicolumn{2}{c|}{$\alpha = 0.4$}                                               & \multicolumn{2}{c|}{$\alpha = 1.0$}                                                  & \multicolumn{2}{c|}{$\alpha = 1.5$}                                             & \multicolumn{2}{c|}{$\alpha = 2.0$}                                                & \multicolumn{2}{c}{Average}                                                  \\
\multirow{-2}{*}{Methods} & reward $\uparrow$                                & cost $\downarrow$                                 & reward $\uparrow$                                & cost $\downarrow$                                 & reward $\uparrow$                                & cost $\downarrow$                                 & reward $\uparrow$                                & cost $\downarrow$                                 & reward $\uparrow$                                & cost $\downarrow$                                 & \multicolumn{1}{c}{reward $\uparrow$}               & \multicolumn{1}{c}{cost $\downarrow$}             \\ \hline
\makecell{ CDT(ours) \\ sampling-based association }  & { \textbf{63.5}} & { \textbf{0.78}} & { \textbf{63.11}} & { \textbf{0.43}} & \textbf{68.13}                         & \textbf{0.31}                        & {\textbf{62.63}}                      & {\textbf{0.68}}                      & { \textbf{58.8}} & { \textbf{0.86}} & { \textbf{63.23}} & { \textbf{0.61}} \\
\makecell{ CDT(ours) \\ density filter }                & { \textbf{62.37}} & { \textbf{0.7}} & { \textbf{62.67}} & { \textbf{0.72}} & \textbf{60.85}                         & \textbf{0.99}                        & {\textbf{67.55}}                      & {\textbf{0.75}}                      & { \color[HTML]{656565} 64.31} & { \color[HTML]{656565} 1.32} & { \textbf{63.39}} & { \textbf{0.76}} \\
\bottomrule % \hline
\end{tabular}
}
\caption{Evaluation results of the normalized reward and cost w.r.t different portions of outlier trajectories. The cost threshold is 1.
$\uparrow$: the higher reward, the better. $\downarrow$: the lower cost (up to the threshold 1), the better. 
Each value is averaged over 20 episodes and 3 seeds.
\textbf{Bold}: Safe agents whose normalized cost is smaller than 1. 
{\color[HTML]{656565} Gray}: Unsafe agents.
}
\label{tab:rebuttal-outliers}
\end{table*}

Apart from the above existing techniques used in our work, we also provide additional ideas to address this issue in different cases, which are inspired by the out-of-distribution (OOD) detection domain. We detail them as follows. 

\begin{itemize}
    \item For the datasets collected in environments with highly stochastic transition dynamics, we can filter outlier trajectories based on the probability of transition dynamics. We first train an empirical transition dynamics density estimator $\hat{p}(s'|s,a): \mathcal{S} \times \mathcal{A} \times \mathcal{S} \rightarrow [0, 1]$ by randomly sample transitions $(s, a, s')$ from the dataset, and then compute the transition probability of each trajectory $\tau$ in the dataset: $p(\tau) = \prod_{t\geq 0} \hat{p}(s_{t+1}|s_t,a_t)$. We can then discard $\alpha \%$ trajectories with the lowest probabilities, where $\alpha$ is the percent of outliers since they are rare in the datasets.
    \item For the datasets that might contain lucky trajectories with high reward and low cost, we can reject the paired Pareto trajectory based on the counts of associated augmentation samples. More specifically, after sampling reward and cost return pairs and finishing association, we can count the number of associated return pairs for each Pareto trajectory. If the count is of a significantly high portion among the total samples, we could discard this Pareto trajectory and continue the process. For example, if we have 100 augmented return pairs in total, and one Pareto data has 80 associated pairs, then we could regard the data as an outlier and remove it from the dataset.
\end{itemize}

Nevertheless, for extremely noisy datasets, the augmentation technique proposed in CDT may still fail.
Investigating how to pre-process the datasets and detect these abnormal trajectories would be interesting for future work.

\clearpage
\subsection{Comparison with more behavior-cloning variants}
\label{app:more-bc}

\begin{figure}[ht]
% \vspace*{-2mm}
\centering
\includegraphics[width=0.95\linewidth]{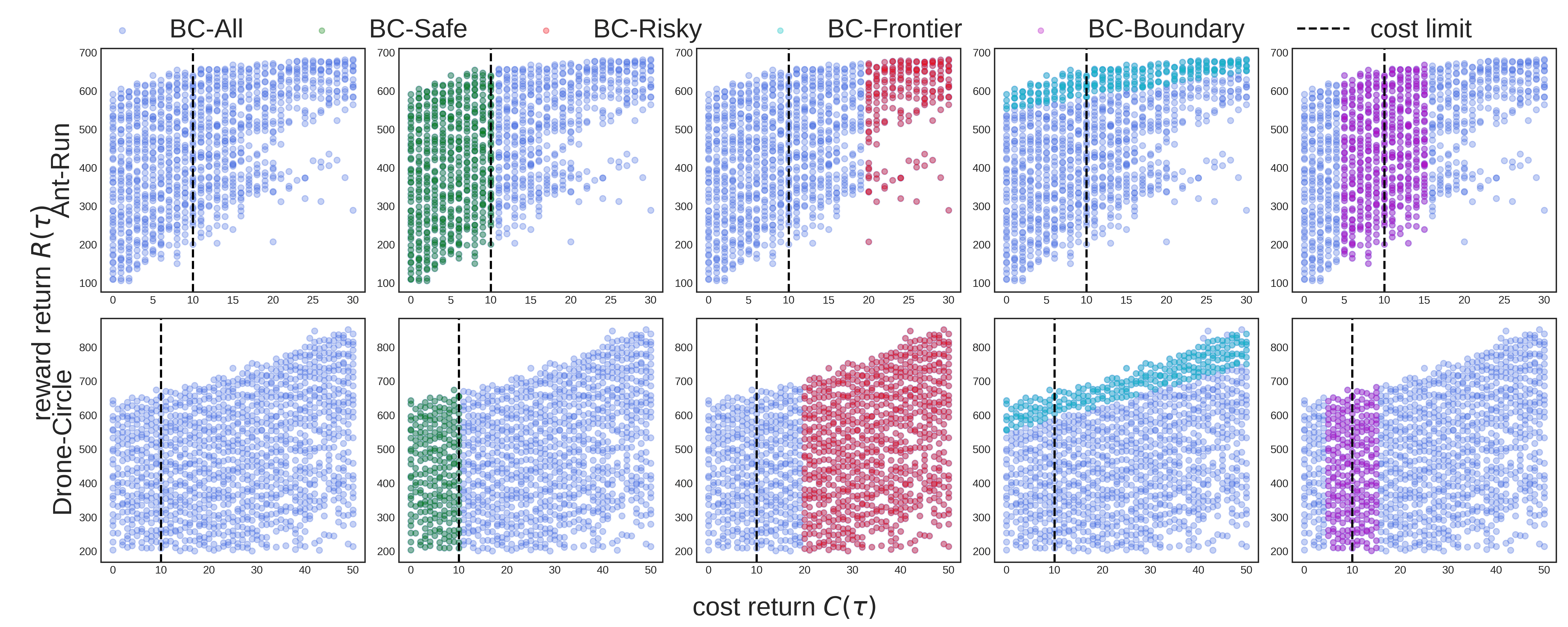}
\vspace*{-2mm}
\caption{Cost-reward return illustrations of the datasets used to train different BC agents (Ant-Run and Drone-Circle dataset).  Each point denotes the trajectories with corresponding cost-return values.}
\label{fig:bc-variants}
% \vspace*{-4mm}
\end{figure}

To determine whether our approach truly employs efficient RL or merely duplicates the offline data, we adopt multiple variants of BC that utilize different portions of the offline data for training agents.
As illustrated in Figure~\ref{fig:bc-variants}, \textbf{BC-all} utilizes the entire set of data, \textbf{BC-Safe} solely relies on safe trajectories, \textbf{BC-risky} exclusively employs high-cost trajectories, \textbf{BC-frontier} utilizes the trajectories that are close to the Pareto frontier, while \textbf{BC-boundary} focuses on the trajectories that are near the cost threshold.

From Table \ref{tab:rebuttal-bc-variants}, we can find that only our approach (CDT) and BC-Safe can successfully learn safe policies for all tasks, with CDT consistently achieving higher rewards than BC-Safe. The under-performance of BC-Risky is not surprising, given that it exclusively utilizes unsafe data.
As expected, BC-risky fails to learn safe policies since it only uses unsafe data.
The poor performance of BC-Boundary and BC-Frontier indicates that relying on a more aggressive expert who can generate high-risk, high-reward trajectories is insufficient for exploring safe boundaries.
BC-All can be deemed an average of the other BC variants and demonstrates the ability to learn safe policies for some tasks.
The comparisons between BC variants and CDT indicates that CDT performs effective RL rather than copying data.

\begin{table*}[ht]
\centering
\renewcommand{\arraystretch}{1.1}
\resizebox{1.\linewidth}{!}{
% \begin{tabular}{c|cc|cc|cc|cc|cc}
\begin{tabular}{c|cc|cc|cc|cc|cc|cc}
% \hline
\toprule
                          & \multicolumn{2}{c|}{Ant-Run}                                                  & \multicolumn{2}{c|}{Car-Circle}                                               & \multicolumn{2}{c|}{Car-Run}                                                  & \multicolumn{2}{c|}{Drone-Circle}                                             & \multicolumn{2}{c|}{Drone-Run}                                                & \multicolumn{2}{c}{Average}                                                  \\
\multirow{-2}{*}{Methods} & reward $\uparrow$                                & cost $\downarrow$                                 & reward $\uparrow$                                & cost $\downarrow$                                 & reward $\uparrow$                                & cost $\downarrow$                                 & reward $\uparrow$                                & cost $\downarrow$                                 & reward $\uparrow$                                & cost $\downarrow$                                 & \multicolumn{1}{c}{reward $\uparrow$}               & \multicolumn{1}{c}{cost $\downarrow$}             \\ \hline
CDT(ours)   & {\color[HTML]{0000FF} \textbf{89.76}} & {\color[HTML]{0000FF} \textbf{0.83}} & {\color[HTML]{0000FF} \textbf{89.53}} & {\color[HTML]{0000FF} \textbf{0.85}} & {\color[HTML]{0000FF} \textbf{99.0}} & {\color[HTML]{0000FF} \textbf{0.45}} & {\color[HTML]{0000FF} \textbf{73.01}} & {\color[HTML]{0000FF} \textbf{0.88}} & {\color[HTML]{0000FF} \textbf{63.64}} & {\color[HTML]{0000FF} \textbf{0.58}} & {\color[HTML]{0000FF} \textbf{82.99}} & {\color[HTML]{0000FF} \textbf{0.72}} \\
BC-Safe     & \textbf{80.56}                        & \textbf{0.64}                        & \textbf{78.21}                        & \textbf{0.74}                        & \textbf{97.21}                       & \textbf{0.01}                        & \textbf{66.49}                        & \textbf{0.56}                        & \textbf{32.73}                        & \textbf{0.0}                         & \textbf{71.04}                        & \textbf{0.39}                        \\
BC-all      & {\color[HTML]{656565} 90.86}          & {\color[HTML]{656565} 1.45}          & \textbf{82.81}                        & \textbf{0.62}                        & \textbf{97.48}                       & \textbf{0.01}                        & {\color[HTML]{656565} 73.29}          & {\color[HTML]{656565} 2.81}          & \textbf{49.58}                        & \textbf{0.28}                        & {\color[HTML]{656565} 78.8}           & {\color[HTML]{656565} 1.03}          \\
BC-risky    & {\color[HTML]{656565} 95.31}          & {\color[HTML]{656565} 3.14}          & {\color[HTML]{656565} 84.1}           & {\color[HTML]{656565} 2.52}          & {\color[HTML]{656565} 96.73}         & {\color[HTML]{656565} 2.71}          & {\color[HTML]{656565} 79.68}          & {\color[HTML]{656565} 3.89}          & {\color[HTML]{656565} 66.74}          & {\color[HTML]{656565} 4.17}          & {\color[HTML]{656565} 84.51}          & {\color[HTML]{656565} 3.29}          \\
BC-boundary & {\color[HTML]{656565} 86.01}          & {\color[HTML]{656565} 1.04}          & \textbf{83.57}                        & \textbf{0.86}                        & \textbf{97.76}                       & \textbf{0.0}                         & \textbf{67.07}                        & \textbf{0.24}                        & {\color[HTML]{656565} 62.93}          & {\color[HTML]{656565} 3.57}          & {\color[HTML]{656565} 79.47}          & {\color[HTML]{656565} 1.14}          \\
BC-frontier & {\color[HTML]{656565} 95.08}          & {\color[HTML]{656565} 1.55}          & {\color[HTML]{656565} 89.76}          & {\color[HTML]{656565} 1.51}          & {\color[HTML]{656565} 98.74}         & {\color[HTML]{656565} 1.58}          & {\color[HTML]{656565} 85.62}          & {\color[HTML]{656565} 3.11}          & {\color[HTML]{656565} 75.36}          & {\color[HTML]{656565} 3.44}          & {\color[HTML]{656565} 88.91}          & {\color[HTML]{656565} 2.24}         \\
\bottomrule % \hline
\end{tabular}
}
\caption{Evaluation results of the normalized reward and cost. The cost threshold is 1.
$\uparrow$: the higher reward, the better. $\downarrow$: the lower cost (up to the threshold 1), the better.
\textbf{Bold}: Safe agents whose normalized cost is smaller than 1. 
{\color[HTML]{656565} Gray}: Unsafe agents.
{\color[HTML]{0000FF} \textbf{Blue}}: Safe agent with the highest reward.
}
\label{tab:rebuttal-bc-variants}
\end{table*}

% To demonstrate the effectiveness of our method in learning a safe and high rewarding policy, we carried out supplementary experiments in more challenging environments in \texttt{safety-gym}~\cite{ray2019benchmarking} including \texttt{Car-Button1}, \texttt{Car-Button2}, \texttt{Point-Button1} and \texttt{Point-Button2}. 
% In the \texttt{Button} environment, a robot must press a goal button while avoiding hazards and gremlins, and while not pressing any of the wrong buttons.
% We applied the same procedure discussed in Appendix \ref{app:data-collection} to prepare the offline datasets.
% Detailed evaluation results of our method and the baselines are presented in Table~\ref{tab:rebuttal-safetygym-exp}. 
% It is evident from our findings that our approach (CDT) is the sole method capable of successfully learning safe policies for all the assigned tasks. 
% The Lagrangian-based baselines, namely BCQ-Lag and BEAR-Lag, have shown a tendency to behave unsafely on the majority of the tasks, implying that directly implementing commonly used safe RL techniques in the offline environment is unlikely to yield favorable outcomes. 
% Additionally, both CPQ and COptiDICE methods, which were specifically designed for offline safe RL, have also struggled to meet the constraints for these challenging tasks.

% \input{tables/rebuttal_safetygym_exp}

\end{document}